\documentclass[11pt]{preprint}

\usepackage[utf8]{inputenc}
\usepackage[full]{textcomp}
\usepackage[osf]{newtxtext} 
\usepackage[title,titletoc]{appendix}

\usepackage{graphicx}
\usepackage{caption}
\usepackage{subcaption}
\usepackage{float}

\usepackage{ifthen} 
\usepackage{hyperref}
\usepackage{breakurl}
\usepackage{amsmath, bm}
\usepackage{amssymb}
\usepackage{times}
\usepackage{amsfonts}
\usepackage{mhequ}
\usepackage{mhenvs}
\usepackage{mhsymb}
\usepackage{mathrsfs}
\usepackage{microtype}
\usepackage{wasysym}
\usepackage{centernot}
\usepackage{booktabs}
\usepackage{tikz}
\usetikzlibrary{snakes}
\usepackage{colortbl}
\usepackage{enumitem}
\usepackage{scalerel}
\usepackage{longtable}
\usepackage{comment}
\usepackage{multirow}

\newtheorem{algorithm}{Algorithm}
\newtheorem{problem}{Problem}
\newtheorem{example}[lemma]{Example}

\def\R{\mathbb{R}}
\def\T{\mathbb{T}}
\def\Z{\mathbb{Z}}
\def\N{\mathbb{N}}
\def\CM{\mathcal{M}}
\def\CO{\mathcal{O}}
\def\CJ{\mathcal{J}}
\def\mcI{\mathcal{I}}
\def\obs{\mathrm{obs}}
\def\pr{\mathrm{pr}}
\def\deg{\mathrm{deg}}

\def\bone{\boldsymbol{1}}
\newcommand{\mrd}{\mathop{}\!\mathrm{d}}

\colorlet{darkblue}{blue!90!black}
\colorlet{darkred}{red!90!black}
\colorlet{darkgreen}{green!70!black}

\makeatletter
\def\blfootnote{\xdef\@thefnmark{}\@footnotetext}
\makeatother

\begin{document}
	
	\date{\today}
	\title{Feature Engineering with Regularity Structures}
	\author{Ilya Chevyrev$^1$, Andris~Gerasimovi\v cs$^2$, and Hendrik~Weber$^3$}
	\institute{The University of Edinburgh, United Kingdom, \email{ichevyrev@gmail.com} \and University of Bath, United Kingdom, \email{andrisger@gmail.com} \and University of M{\"u}nster, Germany,  \email{hendrik.weber@uni-muenster.de}}
	\date{}

	\titleindent=0.65cm
	
	\maketitle
	
	\begin{abstract}
We investigate the use of models from the theory of regularity structures as features in machine learning tasks. A model is a polynomial function of a space-time signal designed to well-approximate solutions to partial differential equations (PDEs), even in low regularity regimes. Models can be seen as natural multi-dimensional generalisations of signatures of paths; our work therefore aims to extend the recent use of signatures in data science beyond the context of time-ordered data. We provide a flexible definition of a model feature vector associated to a space-time signal, along with two algorithms which illustrate ways in which these features can be combined with linear regression. We apply these algorithms in several numerical experiments designed to learn solutions to PDEs with a given forcing and boundary data. Our experiments include semi-linear parabolic and wave equations with forcing, and Burgers' equation with no forcing. We find an advantage in favour of our algorithms when compared to several alternative methods. Additionally, in the experiment with Burgers' equation, we find non-trivial predictive power when noise is added to the observations.
	\end{abstract}

\blfootnote{{\textbf{Funding.} AG and HW were supported by the Leverhulme Trust through a Philip Leverhulme Prize during the writing of this article. HW was also supported by the Royal Society through the University Research Fellowship UF140187.}}
	
	\tableofcontents	
	
	\section{Introduction}\label{sec:intro}
	
	The aim of this paper is to explore the effectiveness of models from Hairer's theory of regularity structures~\cite{Regularity} as feature sets of space-time signals.
	A model is a collection of polynomial functions of the signal which has been used to great success in the analysis of stochastic partial differential equations (SPDEs).
	This paper is the first, to our knowledge, to explore its effectiveness in a machine learning context.

	One of the motivations for this study comes from the fact that models are a higher-dimensional analogue of the \textit{path signature}, a central object in Lyons' theory of rough paths~\cite{lyons1998differential}.
	The signature is the collection of the iterated integrals of a path, which has a rich mathematical structure; it is known to characterise the path up to a natural equivalence relation~\cite{chen_uniqueness,
hambly2010uniqueness,
BOEDIHARDJO2016720}
and leads to a natural notion of non-commutative moments on pathspace~\cite{CL16,CO18Sigs_published}.
	Over the past decade,
	the ability of the signature to encode information about a path in an efficient and robust way has made it a powerful tool in the analysis of \textit{time-ordered data}.
	Examples of
applications of signatures include the recognition of handwriting~\cite{char2,graham2013sparse} and gestures~\cite{gesture1},
	analysis of financial data~\cite{finance1,finance2},
statistical inference of SDEs~\cite{par_est},
	analysis of psychiatric and physiological data~\cite{psych1,sigs_sepsis},
	topological data analysis \cite{CNO20},
	neural networks~\cite{DeepSigs}, and kernel learning~\cite{KO19,CO18Sigs_published}.\footnote{\cite{graham2013sparse,sigs_sepsis} notably received first prizes in the ICDAR 2013 competition and the PhysioNet 2019 Computing in Cardiology Challenge respectively.}
	See~\cite{chevyrev2016primer} for a gentle introduction to the path signature
	and some of its early applications.

\subsubsection*{Our contribution}

Our main contribution is to introduce a novel concept of \textit{model feature vector}  (MFV) that provides an extension of path signatures outside the context of time-ordered data, that is, to data parameterised by \textit{multi-dimensional} space.
The MFV is a collection of space-time functions from $D$ to $\R$, where $D\subset \R^d$ for $d\geq 0$, that is built from an input signal $\xi \colon D \to \R^K$.
In the context of solving PDEs, the signal may incorporate a forcing term and boundary data.
The motivation for the MFV is the fact that solutions to PDEs with a given forcing term and boundary data 
should be well-approximated at a space-time point $z$ by components of the corresponding MFV evaluated at the same space-time point; we give further details in Section~\ref{sec:motivation_SPDE}.
In a machine learning context, the MFV provides a set of features of the original data that can be used in learning algorithms.

In addition to proposing the MFV, we give evidence that these features can carry important information in several test cases.
The basic problem on which we test the use of MFV is:
\begin{problem}\label{prob:1}
For a point $z\in\R^d$, predict the value $u(z)$, where $u$ solves a PDE with a known forcing $\xi$ and boundary condition $u_0$ but with \textit{unknown} coefficients.
\end{problem}
We focus on the case that the PDE in question is an evolution equation.
To address Problem~\ref{prob:1}, we propose two algorithms, Algorithms~\ref{alg:1} and~\ref{alg:2}, in Section~\ref{subsec:algos}
based on elementary linear regression with MFVs in supervised learning tasks.
Algorithm~\ref{alg:1} is designed to predict $u(z)$ in the presence of a general forcing $\xi$,
while Algorithm~\ref{alg:2} is designed to work when there is no forcing (or equivalently $\xi= 0$)
in which case one can leverage the flow property of $u$ to improve predictability.
An important feature of Algorithm~\ref{alg:2} is that it predicts $u(t,x)$ for all space-time points $(t,x)$, thereby effectively learning the entire function $u$.

We investigate the effectiveness of Algorithms~\ref{alg:1} and~\ref{alg:2} in numerical experiments in
Section~\ref{sec:Numerics}.
We apply Algorithm~\ref{alg:1} 
to non-linear parabolic and wave equations with forcing and fixed initial conditions,
and apply Algorithm~\ref{alg:2} to Burgers' equation with no forcing
but varying initial condition.

In the case of Burgers' equation, Algorithm~\ref{alg:2} performs similarly to an adaptation of the PDE-FIND algorithm~\cite{PDE-FIND} on noiseless data and data with small noise, and outperforms the latter on data with large noise (see Section~\ref{subsubsec:comparison}).
In the case of a parabolic equation, Algorithm~\ref{alg:1} outperforms some basic
off-the-shelf regression algorithms (SVR, K-Nearest Neighbours, Random forests) applied simply by treating the forcing as a large vector.

We emphasise that the definition of MFV in Section~\ref{subsec:models} and the algorithms in Section~\ref{sec:algorithm} are presented independently of PDEs
and could be applied to learn other functions of the underlying signal, not necessarily the solution of a PDE -- see Section \ref{sec:discussion} for further discussion.

\subsubsection*{Related works}

The MFV is inspired by the notion of a model from~\cite{Regularity}.
The main difference between our definition and that in~\cite{Regularity}
is that we suitably include the boundary data of the signal as part of the model.
The path signature is a special case of MFV (Proposition~\ref{prop:sigs}).

The idea to apply machine and statistical learning methods to find, predict, or study solutions of PDEs
has seen much attention in recent years.
See for example~\cite{NEURIPS2018_d7a84628,
MR3881695,
MR3874585,
MR3991095,MR3847747,PDE-FIND}
and the references therein.
We also mention the works \cite{Liao_Ming_21_BC,pmlr-v145-li22a,Zhang_etal_20_Meshing, MR4402262,MR4188528} that, like ours, treat boundary data in machine learning-based solvers.
Many works in this direction have focused on new design of learning algorithms.
In contrast, our main contribution comes from designing a new set of features which can be used in a range of algorithms.
As such, we expect our approach to complement many existing methods.

Since the appearance of this article, several works have built on MFVs (or related approaches) especially in combination with neural networks. See for example~\cite{2022arXiv220406255H,NEURIPS2022_09116662}.

	\section{Model feature vectors and regression algorithms}\label{sec:algorithm}
	
	In this section we motivate and define the ``model feature vector'' and introduce two algorithms based on models for learning functions of space-time signals.

We denote by $\N = \{0,1,2,\dots\}$ the set of non-negative integers and by $\R$ the set of real numbers. Assume that we are given a spatial domain $D \subset \R^d$ for $d \geq 0$ and a time horizon $T> 0$.
Given a multi-index $a \in \N^d$, we denote $\partial^a = \partial^{a_1}_{1} \dots \partial^{a_d}_{d}$ where 
\begin{equ}
	\partial^{a_i}_{i} := \frac{\partial^{a_i}}{\partial x_i^{a_i}}\,,\quad\text{for $i = 1, \dots, d$.}
\end{equ}
We will also define the order of a multi-index as $|a| := \sum_{i = 1}^{d} a_i$.
Note that $\partial^0 u =u$.
We let $\partial_t$ denote the partial derivative with respect the time parameter $t\in[0,T]$.

	\subsection{Motivation}\label{sec:motivation_SPDE}

	Our motivating problem is to learn the solution of a PDE on $[0,T]\times D$ given by
	\begin{equs}\label{eq:PDE}
		\mathcal{L} u  &= \mu(\{\partial^{a} u\}_{|a|\leq q}) + \sigma(\{\partial^a u\}_{|a|\leq q}) \xi\,,\\
		u(0,x) &= u_0(x)\;, 
	\end{equs}
where  $\mathcal{L}$ is a linear differential operator,
 $u_0 \colon D \to \R$ is the initial condition,
and $\xi\colon [0,T]\times D \to \R$ is a forcing.\footnote{One might need to include other initial information like an initial speed for the case of the wave equation from Section~\ref{sec:Wave}.}
The functions $\mu,\sigma : \R^{1+d+\cdots +d^q} \to \R$ take as arguments the partial derivatives of $u$ up to order $q$ (i.e. the jet of $u$ to level $q$) and are assumed to be smooth and unknown,
while $(\xi,u_0)$ is known.

	Such an equation is often called a \textit{PDE with forcing} $\xi$.
	When $\xi$ is a random function, e.g. space-time white noise, it is also referred to as a \textit{stochastic} PDE (SPDE).
	
 For this discussion, we assume $\CL = \partial_t - \nu\Delta$ is the heat operator with viscosity $\nu>0$, where $\Delta=\sum_{i=1}^d\partial_i^2$ is the Laplacian on $D\subset\R^d$, and $\mu,\sigma$ depend only on $u$ (and not its derivatives). The case when $\mu,\sigma$ depend on $\partial^a u$ with $|a|>0$ as well as the case of other choices of $\CL$, e.g. the wave operator, are left to the reader.
	
%	Consider an abstract SPDE of the form
%	\begin{equs}[eq:PDE]
%		\partial_t u(t,x) &= \nu \Delta u(t,x) + \mu(u(t,x)) + \sigma(u(t,x))\xi(t,x),\quad(t,x) \in [0,T] \times D\\
%		u(0,x) &= u_0(x)\;,
%	\end{equs}
%	where $D \subset \R^d$ is a smooth domain, $T>0$ is a time horizon, $u_0 : D \to \R$ is the initial condition, $\nu > 0$ is the viscosity, functions $\mu,\sigma : \R \to \R$ are smooth and $\xi$ is a forcing term. In this section we are going to assume that the initial condition $u_0$ is fixed and deterministic.\footnote{We will address random initial conditions in Section~\ref{sec:PDE}}  An example of the forcing $\xi$ that we are going to consider is the space-time white noise on $[0,T] \times [0,1]$. 
	
Under good enough assumptions on the functions $\mu,\sigma,u_0$ and the forcing $\xi$, equation~\eqref{eq:PDE} admits a local in time mild solution.\footnote{In the case that $\xi$ is white noise on $[0,T] \times [0,1]$, smoothness of the above functions is enough (see \cite{DPZ}).} 
In particular, Picard's Theorem implies that $u$ is the limit of the following recursive sequence
\begin{equs}[eq:Picard]
	u^{(0)}&= I_c[u_0]\;,\\
	u^{(n+1)}&= I_c[u_0]+ I[\mu(u^{(n)})] + I[\sigma(u^{(n)})\xi]\;,
\end{equs}
where the operators $I$ and $I_c$ are defined by
\begin{equs}\label{eq:SHE}
		\begin{cases}
			(\partial_t - \nu \Delta) I[f]  = f\,,\\
			I[f](0,\cdot) = 0\,,
		\end{cases}
		\begin{cases}
			(\partial_t - \nu \Delta) I_c[g]  = 0\,,\\
			I_c[g](0,\cdot) = g\,,
		\end{cases}		
	\end{equs}
	for functions $f \colon [0,T] \times D \to \R$ and $g \colon D \to \R$, subject to the same boundary conditions as in~\eqref{eq:PDE}.

The idea is now to Taylor expand the function $\mu$ up to $m$ terms and the function $\sigma$ up to $\ell$ terms in the equation for $u^{(n+1)}$. Define $u^{0,m, \ell} = I_c[u_0]$ and recursively set
\begin{equ}[eq:double_Picard]
	u^{n+1,m, \ell} = I_c[u_0] + \sum_{k=0}^{m} \frac{\mu^{(k)}(0)}{k!} I[(u^{n,m,\ell})^k] + \sum_{k = 0}^{\ell} \frac{\sigma^{(k)}(0)}{k!} I[(u^{n,m,\ell})^k\xi]\,.
\end{equ}
Then, heuristically, since Taylor's expansion implies $u^{n,m,\ell} \to u^{(n)}$ as $m,\ell \to \infty$ and since Picard theorem implies $u^{(n)} \to u$ as $n \to \infty$,
we see that $u^{n,m,\ell}$ should be a good candidate for approximating $u$.

It is not difficult to see from~\eqref{eq:double_Picard} that $u^{n,m,\ell}$ is a polynomial function
of $I_c[u_0]$ and $\xi$ that involves iterated integrals (i.e. iterated applications of $I$).
Recalling that the unknowns are $\mu$ and $\sigma$, and thus $\mu^{(k)}$ and $\sigma^{(k)}$ are also unknown, it is sensible to encode as features these polynomials of $I_c[u_0]$ and $\xi$ and learn the solution map $(u_0,\xi)\mapsto u$ via linear regression. 
Our definition of ``model feature vector'' below precisely encodes this collection of polynomials that appear in $u^{n,m,\ell}$ in a more general setting.
%This concludes our heuristic motivation for using the model $\CM^n_\alpha$ as a set of features for linear regression of solution to~\eqref{eq:PDE}.
These polynomials closely resemble models appearing in the theory of regularity structures, see~\cite[Sec.~8]{Regularity}, which is the motivation behind our terminology.

	\subsection{Model feature vectors}\label{subsec:models}

%\begin{remark}\label{rem:grid}
%For simplicity, the objects in this subsection are formulated in the continuum.
%In practice, however, one typically works with a finite approximation
%and would need to adjust the definitions accordingly;
%see, for instance, Sections~\ref{subsubsec:alg2} and~\ref{sec:Numerics} where we work on a grid.
%\end{remark}

%\begin{remark}
%At this stage, $\xi$ and $u^{(i)}$ look identical, but later on they will play different roles - while $\xi$ is a space-time forcing, the $u^{(i)}$ will be derived from an initial condition to a PDE.
%See Example~\ref{ex:ic} for what $u^{(i)}$ is in our experiments below.
%\end{remark}

We now generalise and abstract the polynomial features discussed in the previous subsection.
Fix for the rest of this section a pair $(\{u^{(i)}\}_{i\in \CJ},\xi)$ (an ``observed signal'') where $\CJ$ is a finite index set (possibly empty) and
\begin{equ}[eq:signal]
\xi = (\xi^{(1)},\ldots, \xi^{(K)})\colon [0,T]\times D\to \R^K\;,\quad u^{(i)}\colon [0,T]\times D\to \R\;.
\end{equ}
We call $\xi$ the \textit{forcing}.
The case $d=0$ corresponds to just $\xi\colon[0,T]\to \R^{K}$ and $u^{(i)}\colon[0,T]\to \R$.
In the experiments in Section~\ref{sec:Numerics}, we sometimes let $u$ be fixed, so that the signal is only $\xi$,
and sometimes we fix $\xi$ (essentially taking $\xi\equiv 0$)
so that the signal is only $u^{(i)}$.
One should think of $\{u^{(i)}\}_{i\in\CJ}$ as ``boundary conditions'', like $I_c[u_0]$ in \eqref{eq:Picard}, and where we allow multiple boundary conditions (as needed, e.g. for the wave equation).

Let us fix a linear operator $I$ that maps space-time functions $f: [0,T]\times D \to \R$ to space-time functions $I[f]$. For example, $I[f]$ could be a convolution with some space-time kernel or a solution to some linear PDE with forcing $f$.
%Given a multi-index $a \in \N^d$, we denote by $\partial^a = \partial^{a_1}_{1} \dots \partial^{a_d}_{d}$ where 
%\begin{equ}
%	\partial^{a_i}_{i} := \frac{\partial^{a_i}}{\partial x_i^{a_i}}\,,\quad\text{for $i = 1, \dots, d$.}
%\end{equ}
%We will also define the order of a multi-index as $|a| := \sum_{i = 1}^{d} a_i$.

%and $g \colon D \to \R$ the functions $I[f],I_c[g]\colon[0,T]\times D\to \R$ by
%define
%\begin{equ}[eq:convolution]
%	I[f](t,x) = \int_0^t\int_D I(t-s, x, y) f(s,y) \mrd s \mrd y\;.
%%	\quad I_c[g](t,x) = \int_D I(t, x, y) g(y) \mrd y\;.
%\end{equ}
%Motivated by models from~\cite[Sec.~8]{Regularity} we give the following definition.
	
	\begin{definition}\label{def:model}
	Consider a tuple of non-negative integers $\alpha = (m,\ell,q) \in \N^3$ and $n\in\N$.
	The \textit{model feature set} $\CS^n_\alpha$ is the finite set of formal symbols defined inductively by\footnote{We use the convention $\prod_{i = 1}^0 \tau_i = 1$.
	Furthermore, the product of symbols is commutative, e.g. we identify $\mcI[\Xi D^1\tau_1D^2\tau_2]
	=\mcI[D^2\tau_2\Xi D^1\tau_1]$,
	and multiplication by $1$ is the identity, i.e. $1\tau=\tau$.}
	$\CS^0_\alpha =\CJ$ and
	\begin{equs}
			\CS^n_\alpha = \Big\{\mcI \big[ \Xi^{(k)} \prod_{i = 1}^p D^{a_i} \tau_i \big] \,:\, 1\leq p+1\leq \ell \Big\} &\cup
 \Big\{\mcI \big[ \prod_{i = 1}^p D^{a_i} \tau_i \big] \,:\, 1\leq p\leq m \Big\} 
\\
&\cup 
 \CS^{n-1}_\alpha\;,
%\,:\, \tau_i \in \CS^{n-1}_\alpha\,,a_i \in \N^d, |a_i| \leq 1\,,\, k,p\in\N\,, \, 1\leq k \leq K\, ,\,
%			\,1 \leq p  +1 \leq \ell  \Big\}
%\\
%&\cup
%\Big\{\mcI\big[ \prod_{i = 1}^p D^{a_i} \tau_i \big]\,:\, \tau_i \in \CS^{n-1}_\alpha\,,a_i \in \N^d, |a_i| \leq 1\,,\, p\in\N\,, 
%			\,1 \leq p  \leq m  \Big\}  \label{eq:model}
%\\
%&\cup \CS^{n-1}_\alpha \,.
		\end{equs}
%		Assume we are given an input $(\{u^{(i)}\}_{i\in\CJ}, \xi)$ of functions \begin{equ}
%\xi, u^{(i)} \colon [0,T]\times D \to \R\;.
%\end{equ}
		% and $\xi \colon [0,T] \times D \to \R$ and 
where $a_i\in\N^d$ with $|a_i|\leq q$, $1\leq k\leq K$, and $\tau_i\in\CS^{n-1}_\alpha$.
Here $\Xi^{(k)}$ and $D^{a_i}$ are formal symbols.
When $a_i=0$ we simply write $D^{a_i}\tau=\tau$.
%(The motivation behind $p+1$ is that $\ell$ is the maximum degree of a monomial containing $\Xi^{(k)}$.)

		The \textit{model feature vector} (MFV, or simply \textit{model}) $\CM^n_\alpha$ of $(u^{(i)},\xi)$ as in \eqref{eq:signal} is the family functions $\CM^n_\alpha\colon \CS^n_\alpha \to \R^{[0,T]\times D}$ that we denote by
		\begin{equ}
		\CM^n_\alpha =(f_\tau)_{\tau\in\CS^n_\alpha}\;,
		\end{equ}
		where $f_\tau\colon [0,T]\times D\to \R$ is defined recursively by $f_i= u^{(i)}$ for $i\in \CJ$ and for $\tau = \mcI\big[\Xi^{(k)}\prod_{i = 1}^p D^{a_i} \tau_i\big]$
and $\sigma = \mcI\big[\prod_{i = 1}^p D^{a_i} \sigma_i\big]$
		\begin{equ}\label{eq:f}
		f_\tau= I\big[\xi^{(k)}\prod_{i = 1}^k \partial^{a_i} f_{\tau_i}\big]\;,\qquad 
f_\sigma= I\big[\prod_{i = 1}^k \partial^{a_i} f_{\sigma_i}\big]\;.
		\end{equ}
%			Observe that, on the right-hand size of~\eqref{eq:f}, $I$, $\partial^{a_i}$ and $\xi$
%are interpreted as linear operators and a function respectively -- in contrast, in the expression for $\tau$, they are interpreted as formal symbols.
		
%		is a family of functions of height $n$, additive width $m$, and multiplicative width $\ell$ is a collection of space-time functions on $[0,T]\times D$ defined recursively by\footnote{We use the convention $\prod_{i = 1}^0 \tau_i = 1$.}
%		\begin{equs}
%			\CM^0_\alpha & = \{I_c[u_0]\}\,,\\
%			\CM^n_\alpha & = \Big\{I\big[\prod_{i = 1}^p \partial^i \tau_i\big],\, I\big[ \xi \prod_{i = 1}^q \partial^ i\tau_i \big]\,:\, \tau_i \in \CM^{n-1}_\alpha\,, \partial^i \in \{\id, \partial_1, \dots, \partial_d \}\,,\label{eq:model}
%			\\
%			&\qquad 1 \leq p \leq m, 0 \leq q \leq \ell \Big\} \cup \CM^{n-1}_\alpha \,,
%		\end{equs}
%		where $\id$ is the identity map and $\partial_i = \frac{\partial}{\partial x_i}$ is the derivative with respect to the $i$-th spatial variable.
		
		We call $n$ the \textit{height} of a model, $m$ the \textit{additive width}, $\ell$ the \textit{multiplicative width}, and $q$ the \textit{differentiation order}.
		Furthermore,
		\begin{itemize}
		\item if $q=0$, we call $\CM^n_\alpha$ a \textit{model without derivatives},
		
		\item if $\ell = 0$, we call $\CM^n_\alpha$ a \textit{model without forcing}, and
		
		\item if $\CJ$ is empty, we call $\CM^n_\alpha$ a \textit{model without initial conditions}.
\end{itemize}
	\end{definition}

We will often use an abuse of notation and write $f_\tau \in \CM^n_\alpha$ meaning that there exists a symbol $\tau \in \CS^n_\alpha$ such that $\CM^n_\alpha[\tau] = f_\tau$.
The symbols in $\CS^n_\alpha$ can be represented as decorated combinatorial rooted trees as in~\cite[Sec.~2]{BHZ}.

Note that additive width $m$ limits how many functions could be multiplied if none of them includes a component of the forcing $\xi$, while multiplicative width $\ell$ limits how many functions could be multiplied if one of them is a component of $\xi$.

\begin{example}
Consider $n = 1$, and $\alpha = (2, 2, 1)$,  and $d = 1$. Suppose $\CJ=\{c\}$ is a singleton.
Then, denoting $D_x = D^{(1)}$,
\begin{equs}
	\CS^{1}_\alpha = \big\{c, \CI[\Xi], \CI[c], \CI[(c)^2], \CI[\Xi c], \CI[D_x c],
	 \CI[cD_x c], \CI[(D_x c)^2], \CI[\Xi D_x c]  \big\}\,.
\end{equs}
If we instead take $\bar \alpha = (2,2,0)$, i.e., consider the model without derivatives, then
\begin{equs}
	\CS^{1}_{\bar \alpha} = \big\{ c, \CI[\xi], \CI[c], \CI[(c)^2], \CI[\xi c] \big\}\,.
\end{equs}
Suppose now that $D=[0,1]$. Let $\xi,u^{(c)}\colon [0,T]\times D \to \R$ be given by $\xi(t,x)=\sin(t)$ and $u^{(c)}(t,x)=\cos(x)$.
Finally, suppose $\CI[f](t,x) = \int_0^x f(t,y)\mrd y$ is the integration-in-space operator.
Then the MFV $\CM^{1}_{\bar \alpha}$ is
\begin{equs}
	\CM^{1}_{\bar \alpha} = \big\{ \cos(x), x\sin(t), \sin(x), (x+\sin(x)\cos(x))/2, \sin(t) \sin(x) \big\}\,,
\end{equs}
where we used $\int_0^x\cos(y)\mrd y=\sin(x)$ and $\int_0^x \cos(y)^2\mrd y = (x+\sin(x)\cos(x))/2$.
The space-time functions $f_\tau$ in $\CM^{1}_{\bar \alpha}$ correspond to the symbols of $\CS^{1}_{\bar \alpha}$ in the same order, e.g., $f_{\CI[(c)^2]} = (x+\sin(x)\cos(x))/2$.

To give an example at level $n=2$, one of the symbols in $S^2_\alpha$ is $\tau=\CI[\Xi \CI[cD_x c]]$.
The corresponding function $f_\tau \in \CM^2_\alpha$ is
\begin{equs}
f_\tau(t,x)
&= \int_0^x \sin(t) \Big(\int_0^y \cos(z)(-\sin(z)) \mrd z\Big) \mrd y
= \int_0^x \sin(t)\frac12(-\sin^2(y)) \mrd y 
\\
&= \frac18 \sin(t) (\sin(2x)-2x)\;.
\end{equs}
\end{example}

We next show precisely how the path signature is generalised by the MFV.
Consider $n\geq 1$ and a differentiable path
\begin{equ}
X=(X^{(1)},\ldots, X^{(K)}) \colon [0,T]\to \R^K\;.
\end{equ}

\begin{definition}\label{def:sigs}
The level-$n$ \textit{signature} of $X$ over an interval $[s,t]\subset[0,T]$ is the collection of $K^n$ numbers $\{S^I_{s,t}(X)\}_{I}$ indexed by multi-indexes $I=(i_1,\ldots,i_n) \in \{1,\ldots,K\}^n$
and defined by the iterated integrals
\begin{equ}[eq:sig_def]
S_{s,t}^{(i_1,\ldots, i_n)}(X) = \int_s^t \int_s^{t_{n}} \ldots \int_s^{t_{2}} \dot X^{(i_1)}_{t_1} \ldots \dot X^{(i_{n-1})}_{t_{n-1}}\dot X^{(i_n)}_{t_{n}} \mrd t_1 \ldots \mrd t_{n-1}\mrd t_n\;.
\end{equ}
\end{definition}

\begin{proposition}\label{prop:sigs}
Define the family of symbols $\CW^n$ inductively by $\CW^0 = \{1\}$ and
\begin{equ}
\CW^n = \{\CI [\Xi^{(k)} \tau  ] \,:\, \tau \in \CW^{n-1}\,,\; 1\leq k\leq K\}\;,
\end{equ}
where $\Xi^{(k)}1:= \Xi^{(k)}$.
Then there is a bijection $\phi \colon \{1,\ldots,K\}^n\to \CW^n$ given by
\begin{equ}
\phi(i_1,\ldots,i_n) = \CI[\Xi^{(i_n)}\CI[\Xi^{(i_{n-1})}\CI[\ldots \CI[\Xi^{(i_1)}]\ldots] ]]\;.
\end{equ}
Consider furthermore $d=0$ and define the $\R^K$-valued forcing \begin{equ}
\xi=\{\xi^{(i)}\}_{i=1}^K\colon[0,T]\to \R^K\;\qquad \xi^{(i)}=\dot X^{(i)}\;.
\end{equ}
Let $I[\xi](t)=\int_0^t \xi(s)\mrd s$ be the integration-in-time operator
and $\alpha = (0,2,0)$ and $\CJ = \emptyset$
(i.e. consider the model without initial conditions).

Then $\CW^n \subset \CS^n_\alpha$
and, for all $(i_1,\ldots,i_n)\in \{1,\ldots,K\}^n$,
\begin{equ}[eq:model_sig]
\CM^n_\alpha [\phi(i_1,\ldots,i_n)] (T)= S_{0,T}^{(i_1,\ldots, i_n)}(X)\;.
\end{equ}
\end{proposition}

\begin{proof}
The inclusion $\CW^n \subset \CS^n_\alpha$ and that $\phi \colon \{1,\ldots,K\}^n\to \CW^n$  is a bijection are clear.
Equality \eqref{eq:model_sig} is clearly true for $n=1$ and in general follows by induction:
\begin{equs}
\CM^n_\alpha [\phi(i_1,\ldots,i_n)](T)
&= \CM^n_\alpha [\CI[\Xi^{(i_n)} \phi(i_1,\ldots,i_{n-1})]](T)
\\
&= \int_0^T \dot X^{(i_n)}  S^{(i_1,\ldots i_{n-1})}_{0,t_n}(X) \mrd t_n
\\
&= S^{(i_1,\ldots i_n)}_{0,T}(X)
\end{equs}
where the second equality follows from the inductive hypothesis of \eqref{eq:model_sig} for $n-1$
and the third equality follows readily from the definition \eqref{eq:sig_def}.
\end{proof}

In our experiments below, the finite index set $\CJ$ and ``boundary conditions'' $\{u^{(i)}\}_{i\in\CJ}$ will be taken as follows:
in Section~\ref{sec:numerics_Burgers} $\CJ$ will be a singleton $\CJ=\{c\}$ and $u^{(c)}$ will be the solution to the linear heat equation with a given initial condition $u_0^{(c)}\colon D\to \R$, i.e. $u^{(c)} = I_c[u_0^{(c)}]$ for $I_c$ as in Section \ref{sec:motivation_SPDE};
in Section~\ref{sec:Parabolic} $\CJ$ will primarily be empty $\CJ=\emptyset$ as we will ignore initial conditions (though see Section \ref{subsubsec:2D} for an exception);
in Section~\ref{sec:Wave} $\CJ$ will contain two elements $\CJ=\{c,s\}$ 
and $u^{(c)}$ (resp. $u^{(s)}$) will be the solution of the linear wave equation with initial condition $u^{(c)}_0$ and initial speed $0$ (resp. initial condition $0$ and initial speed $u^{(s)}_0$), where both $u^{(c)}_0, u^{(s)}_0$ are given.
%see Definition~\ref{def:model_wave}).

%	In words, the model $\CM^n_\alpha$ contains functions that were obtained from $u^i$ by iteratively taking partial derivatives, multiplications between each other or with the forcing $\xi$, and convolution with the kernel $I$. 

While we consider only a space-time setting,
	Definition~\ref{def:model} readily adapts to
an purely spatial setting. In this case the linear operator $I$ would map functions $f\in \R^D$ to $I[f]\in \R^D$, e.g. $I[f](x)=\int_D K(x,y)f(y)\mrd y$ for a kernel $K\colon D\times D\to \R$.

\subsection{Regression algorithms}
\label{subsec:algos}

In this subsection, we propose two supervised learning algorithms which use the MFV of an input $(\{u^{(i)}\}_{i\in \CJ},\xi)$ to learn an output $u$.
While in principle there is no limitation of the nature of $u$ (vector, classification label, etc.), we will consider the special case where
$u$ is a number associated to a space-time point or is a space-time function.
In the experiments in Section~\ref{sec:Numerics}, $u$ will be the solution to a PDE with forcing $\xi$ and a given initial condition.
We will furthermore consider henceforth $K=1$, so the forcing is $\R$-valued and simply write $\xi^{(1)}=\xi$;
the generalisation $K>1$ is left to the reader.

\subsubsection{Prediction at one point}
\label{subsubsec:predict_one_point}

In the following algorithm, one should think of the observation $u$ as a quantity which depends on the signal $(\{u^{(i)}\}_{i\in \CJ},\xi)$ at a given space-time point $(t,x)\in[0,T]\times D$.
Below $\{u^{(i)}\}_{i\in \CJ}$ and $\xi$ will denote
functions $u^{(i)},\xi\colon[0,T]\times D \to \R$ for every $i\in \CJ$.
	
	\begin{algorithm}[Prediction at one point.]\label{alg:1}\leavevmode

\textbf{Parameters:} integers $n,m,\ell,q \in \N$ and an operator $I$.

\textbf{Input:}
\vspace{-3.5mm}

\begin{itemize}
\setlength\itemsep{-1mm}
\item a point $(t,x)\in[0,T]\times D$;
\item a set $\CJ$;
\item set of observed triplets $(u,\{u^{(i)}\}_{i\in \CJ},\xi) \in U^\obs$ where $u\in \R$;
\item a set of pairs $(\{v^{(i)}\}_{i\in \CJ},\zeta)\in U^\pr$ for which we want to make a prediction.
\end{itemize}
\vspace{-1.5mm}

\textbf{Output:}  Prediction $u^\pr\in\R$ for every $(\{v^{(i)}\}_{i\in \CJ},\zeta)\in U^\pr$.

		\begin{enumerate}[label=Step~\arabic*]
			\item Let $\alpha = (m,\ell,q)$. For each $(u,\{u^{(i)}\}_{i\in \CJ},\xi) \in U^\obs$ and (resp. each $(\{v^{(i)}\}_{i\in \CJ},\zeta) \in U^\pr$) construct a model $\CM^n_\alpha=(f_\tau)_{\tau\in\CS^n_\alpha}$ using $\{u^{(i)}\}_{i\in \CJ}$ and $\xi$ (resp. $\{v^{(i)}\}_{i\in \CJ}$ and $\zeta$)
			\item\label{alg1:step:2} Fit a linear regression of $u$ against $(f_\tau(t,x))_{\tau \in \CS^n_\alpha}$ for each $(u,\{u^{(i)}\}_{i\in \CJ},\xi) \in U^\obs$.
			\item For each $(\{v^{(i)}\}_{i\in \CJ},\zeta) \in U^\pr$, construct a prediction $u^\pr$ using the linear fit constructed from~\ref{alg1:step:2} and the associated model $\CM^n_\alpha$.
		\end{enumerate}
	\end{algorithm}

Recall that our motivating problem is to learn the solution of a PDE \eqref{eq:PDE} at a given point $(t,x)$ where $(u_0,\xi)$ are observed but $\mu,\sigma$ are unknown.
Using the notation of Section \ref{sec:motivation_SPDE},
our typical choice for $\CJ$ is a singleton $\CJ=\{c\}$
with $u^{(c)} = I_c[u_0]$
(but we use other choices if we wish to encode more or less boundary conditions).
The heuristic reason why Algorithm~\ref{alg:1} should work for predicting PDEs comes from the fact that functions in $\CM^n_\alpha$ constructed from $\xi$ and $I_c[u_0]$
well approximate the $n$-th Picard iterate $u^{(n)}$ which itself should converge to the solution of~\eqref{eq:PDE} for smooth $\mu$ and $\sigma$.
%For the PDE \eqref{eq:PDE}, one would take
%$q \leq 1$ because there could be at most one derivative on the right-hand side of~\eqref{eq:PDE}.
	
\begin{remark}\label{rem:polinoms}
If it is known that the equation~\eqref{eq:PDE} is additive, i.e. that $\sigma$ is a constant, then the heuristic of Section \ref{sec:motivation_SPDE} suggests that one should consider $\CM^n_\alpha$ with $\ell = 1$. More generally, if it is known that both $\mu$ and $\sigma$ are polynomials, then the heuristic suggests that taking $m$ and $\ell-1$ greater than the respective degrees of $\mu$ and $\sigma$ would likely not improve the accuracy of the above algorithm.	
These remarks follow from the fact that polynomials agree with their Taylor expansion (for high enough order of expansion).
\end{remark}

	Note that, in Algorithm~\ref{alg:1}, we regress against the functions in the model at one input space-time point $(t,x)$ only (see Section~\ref{sec:motivation_SPDE} for the motivation behind this choice in the case of PDEs). 
In the case of path signatures (Definition~\ref{def:sigs}), this corresponds to using only the endpoint $T$ of the signature, i.e. $S^{I}_{0,T}(X)$, which is common practice (see e.g.~\cite{psych1, KO19, CNO20}).
There are situations, however, where it is beneficial to
use the signature of a path over different segments, i.e. use $S^I_{s,t}$ as a feature for different choices of $[s,t]\subset[0,T]$, and the choice of
segments is a hyperparameter,
see e.g. the sliding window approach of~\cite{char2}.
It would be of interest to explore if a similar approach yields any benefit for MFVs.

\subsubsection{Prediction using flow property}
\label{subsubsec:alg2}

We will now focus on predicting functions $u$ defined on all space-time points which have a given initial condition and no forcing.
Algorithm~\ref{alg:2} below is designed to work when $u$ satisfies the \textit{time-homogeneous flow property}:
%starting from a function at time $t = 0$ from initial condition $u_0$ and starting function at time $t=s$ from the initial condition $u_s$ should give the same result on the time interval $[s,T]$.
$u(t,x)$ should depend on $u(0,\cdot)$ in the same way as $u(t+h,x)$ depends on $u(h,\cdot)$.

The algorithm employs a discretisation of time $\CO_T = \{0=t_0<t_1<\ldots<t_N=T\}\subset[0,T]$ which we assume is equally spaced, i.e. $t_k = \delta k$ where $\delta = T/N$.
The observed and predicted functions of this algorithm are both functions $\CO_T\times D \to \R$.

%Assume that the observed time points $\CO_T$ are equally distanced. Let $\delta = T/N$ and $t_k = \delta k$ be the $k$-th observed time point. 
%Assume that we are given an additional linear map $I_c$ which is an \textit{initialising map}: given $u_0 \colon \CO_X \to \R$, $I_c[u_0]$ is another function $\{0,\delta\}\times\CO_X \to \R$.

%The main choice for initializing map $I_c$ that we are going to use later on is\footnote{Once again, the integral will be computed using the approximation $\CO_X$.}
%	\begin{equ}[eq:initializing_map]
%		I_c[u_0] (t,x) = \int_D I(t,x,y)u_0(y) dy\;.
%	\end{equ}

Assume further that we are given an additional linear map $I_c$ which is an \textit{initialising map}: given $u_0 \colon D \to \R$, $I_c[u_0]$ is another function $[0,\delta]\times D \to \R$.
Let $\CM^n_\alpha(u_{0})$ be the model without forcing ($\ell = 0$) constructed on $[0,\delta] \times D$ with $\CJ = \{c\}$ and $u^{(c)} := I_c[u_{0}]$.
	
We briefly describe the algorithm in words.
Suppose that we know or have a prediction for $u(t_k,x)$ for some $k \in \{ 0, \dots, N-1\}$ and all $x\in D$.
	Under the time-homogeneous flow property, it is natural to seek an approximation for $u(t_{k+1},x)$ using a functional linear regression of the form
	\begin{equ}\label{eq:approxPDE}
		u(t_{k+1},\cdot) \approx a(\cdot) + \sum_{\tau \in \CS^n_\alpha} b_\tau(\cdot) f_\tau(\delta, \cdot)\,,
	\end{equ}
	where $a,b\colon D\to \R$ are functions to be learned and $f_\tau \in \CM^n_\alpha(u_{t_k})$. 
The time homogenous flow property implies that $a$ and $b_\tau$ are expected to only depend on the time step $\delta$ and not $t_k$ (but $a,b_\tau$ can depend on $x\in D$).
In the training phase, we therefore decompose each observation $u\colon \CO_T\times  D \to\R$ into $N$ `subobservations' $u(t_k,\cdot)\colon D\to\R$, for $k=0,\dots,N-1$,
and learn the coefficients $a,b_\tau$ from these subobservations.
The prediction phase then recursively applies the formula~\eqref{eq:approxPDE} to predict $u$ from the `initial condition' $u(0,\cdot)$.
In the following, we will sometimes write $u_{t}$ for the function $u(t,\cdot)\colon D\to\R$.
	
	\begin{algorithm}[Prediction using flow property.]\label{alg:2} \leavevmode

\textbf{Parameters:} integers $n,m,q \in \N$, operator $I$, and initialising map $I_c$.

\textbf{Input:}
\vspace{-3.5mm}

\begin{itemize}
\setlength\itemsep{-1mm}
\item a collection $\{u(t,x)\}_{(t,x)\in \CO_T\times D}\in U^{\obs}$ of observed functions;

\item a collection $u_0\in U^\pr$  of initial conditions $u_0\colon D\to\R$ for which we want to make a prediction.
\end{itemize}
\vspace{-3.5mm}

\textbf{Output:} A prediction $u^\pr \colon \CO_T\times D\to \R$ for every $u_0\in U^\pr$.

		\begin{enumerate}[label=Step~\arabic*]
			\item Let $\alpha = (m, 0, q)$. For $k = 0, \dots N-1$ and each $u \in U^\obs$ construct a model $\CM^n_\alpha(u_{t_k})$ on $[0,\delta] \times D$ with $\CJ=\{c\}$ and $u^{(c)} = I_c[u_{t_k}]$. 
			
			\item\label{step:2} For each $x \in D$ fit a linear regression as in~\eqref{eq:approxPDE} of
\begin{equ}
(u(t_{j+1}, x))_{u \in U^\obs, j = 0,\dots,N-1}
\end{equ}
against
\begin{equ}
\big((f_{\tau} (\delta, x))_{f_\tau \in \CM^n_\alpha(u_{t_j})}\big)_{u \in U^\obs, j = 0,\dots,N-1}\;.
\end{equ}
			
			\item\label{step:3} For each $u_0 \in U^\pr$ construct a model $\CM^n_\alpha(u_0)$ on $[0,\delta]\times D$ with $\CJ=\{c\}$ and $u^{(c)} = I_c[u_0]$.
			Make a prediction of $u^\pr(t_{1}, x)$ for each $x \in D$ based on the fit from~\ref{step:2} and $(f_\tau(\delta, x))_{f_\tau \in \CM^n_\alpha(u_0)}$. 
			
			\item\label{step:4} Recursive step. For each $u_0 \in U^\pr$, $k \geq 1$, and the predicted $u^\pr_{t_k}$, construct a model $\CM^n_\alpha(u^\pr_{t_k})$ on $[0,\delta] \times D$ with $\CJ=\{c\}$ and $u^{(c)} = I_c[u^\pr_{t_k}]$ and make a prediction of $u^\pr(t_{k+1}, x)$ for each $x \in D$ based on a linear fit from~\ref{step:2} and $(f_\tau(\delta, x))_{f_\tau \in \CM^n_\alpha(u^\pr_{t_k})}$.
		\end{enumerate} 
	\end{algorithm}
	
	When specific boundary values are given, one might need to enforce these for the predicted function $u^\pr$. For example one might set $u^\pr(t_k,x) = 0$ for $x \in \partial D$ and every $k = 0,\dots N$ if this was known.
	
	As remarked earlier, Algorithm~\ref{alg:2} effectively converts the size of the training set for the linear fit from $|U^\obs|$ to $N\times|U^\obs|$.

Algorithm~\ref{alg:2} aims to address a problem similar to that
of learning a dynamical system.
A different approach to this problem is dynamic mode decomposition, which is based on spectral analysis of the Koopman operator~\cite{Schmid10,Koopman}.
	
\subsubsection{Feature selection and hyperparameters}
\label{subsubsec:degree}

	The cardinality of $\CS^n_\alpha$ grows exponentially with $n$. To avoid overfitting or to speed up the learning, it can be important to restrict further the number of elements in $\CS^n_\alpha$.
	We do this below by introducing a function called \textit{degree} $\deg \colon \CS^n_\alpha \to \R$ which satisfies $\deg(\xi)=\eta$ and $\deg(i)=\eta_i$, $i\in\CJ$, for some $\eta,\eta_i\in\R$ together with the inductive definition
	\begin{equ}[eq:degree]
		\deg I[\tau] = \beta + \deg \tau\,,\quad \deg \partial^{a_i} \tau = \deg \tau - |a_i| \,, \quad \deg \prod^k_{i = 1} \partial^{a_i} \tau_i = \prod^k_{i = 1} \deg \partial^{a_i} \tau_i\,,
	\end{equ}
	for some $\beta > 0$.
	This definition is set so that $\deg$ of symbols from $\CS^n_\alpha$ will be usually larger for bigger $n,m,\ell,q$.
	We will then perform the regression in our algorithms against the functions in the model whose symbol does not exceed a certain degree $\gamma$.
	When used, the degree function and cutoff $\gamma$ are additional parameters in Algorithms~\ref{alg:1} and \ref{alg:2}.
	
	We follow a ``rule of thumb'' of keeping the ratio (number of train cases):(number of predictors) above $10$ 
	(see~\cite[Sec.~4.4]{Harrel2015Book} and references therein for a discussion about such rules). Thus, one would choose $\gamma$ so that the number of functions in $f_\tau \in \CM^n_\alpha$ with $\deg\tau \leq \gamma$ is at least $10$ times smaller than number of elements in $U^\obs$.
	
	Note that it is much easier to keep the train cases to predictors ratio above $10$ for Algorithm~\ref{alg:2} because we have $N|U^{\obs}|$ train cases compared to only $|U^{\obs}|$ train cases in Algorithm~\ref{alg:1}. Nevertheless, it is still beneficial to use degree for Algorithm~\ref{alg:2} for computational reasons. Indeed, the linear regression time complexity for Algorithm~\ref{alg:2} is $O(|\CM^n_\alpha|^3+|\CM^n_\alpha| N|U^{\obs}|)$.
Thus, a large size of the model can drastically slow down the learning.

The use of $\deg$ and cutoff $\gamma$ is motivated by analysis of SPDEs,\footnote{The notion of degree is similar to the one introduced in~\cite{Regularity} and is related to the H\"older regularity of functions in the model which are built from highly oscillatory signals.
%		If $I$ is given by a convolution with the heat kernel and $\xi$ and $u_0$ are highly oscillatory signals,
%		then, heuristically, functions with smaller degree are fluctuating more rapidly and therefore might be more important for predictability.
	}  but other choices of feature selection are possible and may lead to improved learning.
For example, it is possible to consider higher degree features but with a sparsity (i.e. $l_0$ norm) penalty, which is often employed in dictionary learning, 
though this choice would still require the computation of a large model $\CM^n_\alpha$, at least on the training data;
it would be of significant interest to investigate this form of feature selection (we are not aware of any systematic studies of sparse dictionary learning even for signature features).

In addition to the degree,
Algorithms \ref{alg:1} and \ref{alg:2} come with several further hyperparameters, one of which is the `height' (number of iterated applications of a linear operator) of the model;
in the case of path signatures (Definition \ref{def:sigs}), our `height' is the `level' of a signature.
In all the numerical experiments in Section \ref{sec:Numerics}, it was established that using a model with a larger height improves the performance of regression.
Another hyperparameter is the linear operator $I$ used in the definition of a model.
In the case of Burgers' equation analysed in Section \ref{sec:numerics_Burgers}, we additionally found that $I$ can be `guessed' from the data, yielding sensible results
(the guess for $I$ does not need to be precise but the precision influences the prediction power).
Some further discussion is given in  Section~\ref{sec:discussion}.

%	\begin{remark}\label{rem:slow_F}
%		If we know that the dependence of $u_{t_{k+1}}$ on $u_{t_k}$ changes slowly and continuously with time, we can modify~\ref{step:2} of Algorithm~\ref{alg:2} and say that for each $k = 0,\dots,N-1$ we regress $(u(t_{j+1},x))_{u \in U^\obs, j \in J_k}$ against $\big((f_\tau(t_{j+1},x))_{f_\tau \in \CM^n_\alpha(u_{t_j})}\big)_{u \in U^\obs, j \in J_k}$ where $J_k = \{0\leq \ell \leq N-1\,:\, |\ell - k| \leq p \}$ for some relatively small $p$. This is motivated by the fact that even though coefficients in~\eqref{eq:approxPDE} now do depend on $t_{k+1}$ they are expected to be not too different for $t_{k-p+1},\dots,t_{k+p+1}$ if $p$ is not large.
%	\end{remark}

	\section{Numerical simulations}\label{sec:Numerics}
	
We present several numerical experiments where we learn the solution of the PDE \eqref{eq:PDE} with different choices of operator $\CL$ and non-linearities $\mu,\sigma$.
	In general one needs to specify the boundary conditions of \eqref{eq:PDE}, i.e. the values of $u(t,x)$ for $x \in \partial D$. For simplicity, we only consider periodic boundary conditions in our experiment, but Dirichlet or Neumann boundary conditions can be easily implemented.
As in Section \ref{subsec:algos}, we will only consider MFVs (Definition \ref{def:model}) with $K=1$ and $q\leq 1$.

To approximate the continuum,
we fix a finite grid $\CO\subset [0,T]\times D$.
We will assume that $\CO = \CO_T \times \CO_X$ where $\CO_T = \{0 = t_0 < t_1 < \dots < t_{N} = T\}$
for some integer $N\geq1$ and $\CO_X$ is a finite grid of points in $D$.
We will work with functions defined on the grid $\CO$ instead of $[0,T]\times D$. For this purpose, the operator $I$, the partial derivatives $\partial_i$, and $I_c$ (whenever it is used) must have approximations on $\CO$.

	In all experiments below we use an ordinary least squares linear regression. See
\begin{center}
\href{https://github.com/andrisger/Feature-Engineering-with-Regularity-Structures.git}{https://github.com/andrisger/Feature-Engineering-with-Regularity-Structures.git} 
\end{center} for Python code containing implementation of the model and experiments from this section.
	
	\subsection{Parabolic PDEs with forcing}\label{sec:Parabolic}
	
In this subsection we will suppose that the differential operator in~\eqref{eq:PDE} is given by $\mathcal{L} = \partial_t - \nu \Delta$, where $\nu > 0$ is the viscosity and $\Delta=\sum_{i=1}^d\partial_i^2$ is the Laplacian on $D\subset\R^d$. This motivates the following definition.
%In this case the kernel $I$ is a heat kernel. For example, the case of periodic boundary conditions on $D=[0,1]^d$, $I(t,x,y)$ will be periodised version of the usual heat kernel $(4\pi \nu t)^{-d/2}\exp(-\|x-y\|^2/4\nu t)$. 
	
	\begin{definition}\label{def:model_heat}
		Fix an initial condition $u_0 \colon D \to \R$ and a forcing $\xi\colon[0,T] \times D \to \R$ as well as $n,m,\ell \geq 1$ and $\nu>0$. Let $q \leq 1$ and $\alpha = (m,\ell,q)$. The model $\CM^n_\alpha$ for the parabolic equation with viscosity $\nu$ is constructed by taking $\CJ = \{c\}$ with $u^{(c)} = I_c[u_0]$ where operators $I$ and $I_c$ are given by \eqref{eq:SHE}.
	\end{definition}

	\begin{remark}\label{rem:viscosity}
		Algorithm~\ref{alg:1} does not require knowledge of $\mu$ or $\sigma$ in~\eqref{eq:PDE}.
		However, in the experiments in this subsection, $\mu$ and $\sigma$ will be polynomials, and we will use knowledge of their degree to choose the hyperparameters $m,\ell$.
		Another hyperparameter is $\nu$ since this determines $I$ through~\eqref{eq:SHE}.
	When $\mu,\sigma,\nu$ are completely unknown, these hyperparameters could be chosen, as usual, by splitting the data into training, validation and test sets, and tuning the hyperparameters on the validation set.
	See also Section~\ref{sec:numerics_Burgers} where a starting point for an approximation of the viscosity $\tilde{\nu}$ is derived from the training data.
	\end{remark}
	
	\subsubsection{Multiplicative forcing}\label{sec:parabolic_multi}
	
	Consider the following PDE
	\begin{equs}[eq:phi4m]
		(\partial_t - \Delta) u &= 3u - u^3 + u\,\xi\,\quad\text{for $(t,x) \in [0,1]\times [0,1]$,}\\
		u(t,0) &= u(t,1) \quad\text{(Periodic BC),}\\
		u(0,x) &= x(1-x)\,,
	\end{equs}
	where $\xi$ is a space-time forcing. Here we discretise space and time respectively in $100$ and $1000$ evenly distanced points, which we use to define the grids $\CO_X$ and $\CO_T$.
	We solve~\eqref{eq:phi4m} for each forcing $\xi$ using a finite difference method on the same discretisation $\CO=\CO_T\times\CO_X$ (see \cite[Sec.~10.5]{numericalSPDE}). 
	
	We take here $\xi$ as approximations of space-time white noise.
	We performed Algorithm~\ref{alg:1} both using the full model $\CM^n_\alpha$ from Definition~\ref{def:model_heat} with viscosity $\nu=1$ and the model without the initial conditions (i.e. where $\CJ$ is assumed to be empty in the construction of $\CS^n_\alpha$). We have found that in practice using the 
	full model did not drastically improve the errors (see Remark~\ref{rem:reduced_model}). Therefore, we primarily present results in this subsection for the model without the initial condition.
	
	We construct a model with $\CJ = \emptyset$ of height $n = 4$, additive width $m = 3$, multiplicative width $\ell=2$,\footnote{See Remark~\ref{rem:polinoms} for a motivation behind taking these particular widths.}
 and differentiation order $q = 0$ (because $\mu$ and $\sigma$ do not depend on $\partial_iu$) so that $\alpha = (3,2,0)$. We assign a degree from Section~\ref{subsubsec:degree} to satisfy~\eqref{eq:degree} with $\beta = 2$ and $\deg \xi = -1.5$.\footnote{This is motivated by the H\"older regularity of space-time white noise being $-1.5 - \varepsilon$ for any small $\varepsilon > 0$ and the fact that the heat operator $I$ increases the H\"older regularity by $2$.} In the experiments below we only consider functions $f_\tau \in \CM^4_{\alpha}$
with $\deg \tau \leq 5$.
	
\begin{figure}[h]
		\begin{subfigure}[h]{0.48\textwidth}
			\centering
			\includegraphics[width=\textwidth]{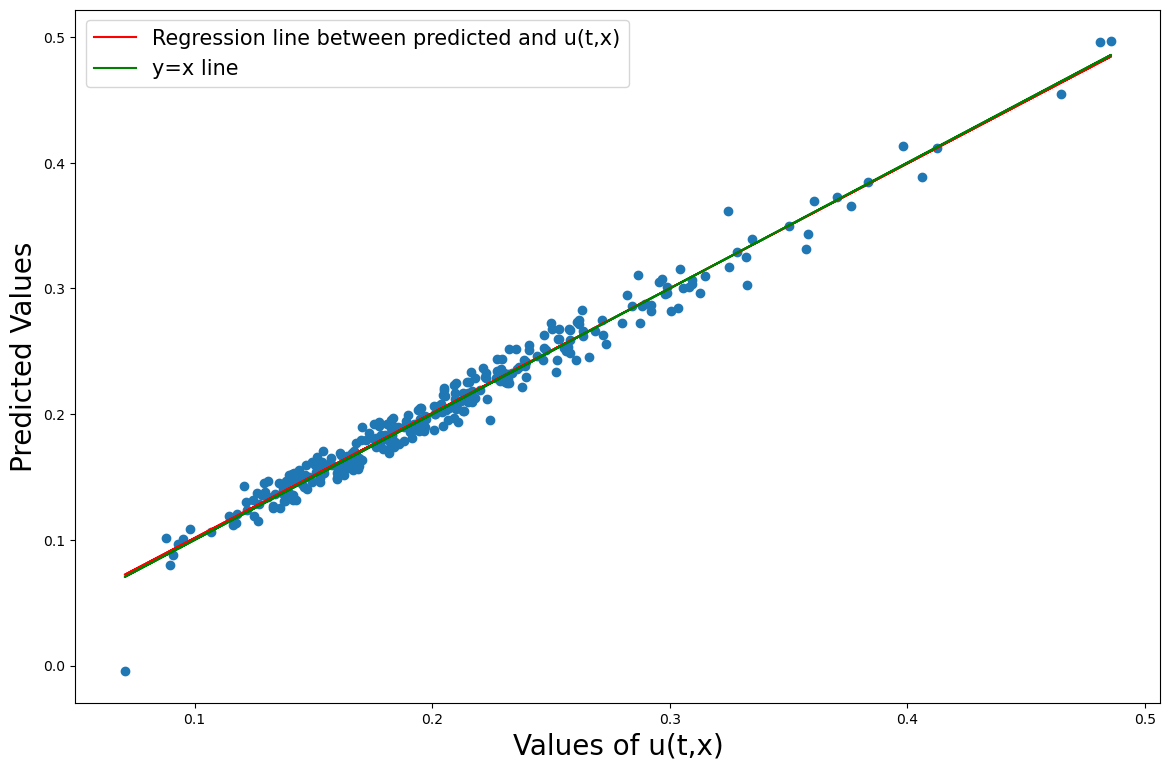}
			\caption{Prediction at $(t,x) = (0.05,0.5)$.\\ Relative $\ell^2$ error: $4.8\%$. Slope: $0.99$. Error standard deviation: $4.1\%$. $R^2 = 0.98$.}
		\end{subfigure}
\hfill
		\begin{subfigure}[h]{0.48\textwidth}
			\centering
			\includegraphics[width=\textwidth]{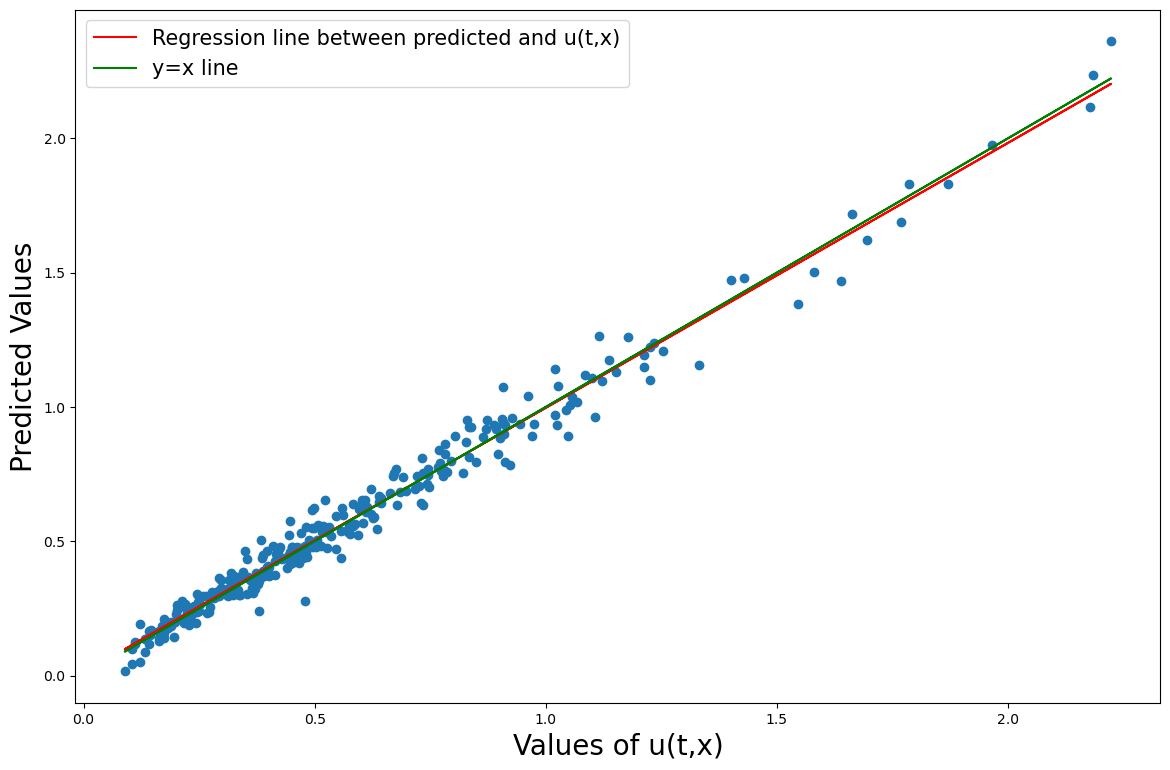}
			\caption{Prediction at $(t,x) = (0.5,0.5)$.\\ Relative $\ell^2$ error: $7.8\%$. Slope: $0.99$. Error standard deviation: $5.2\%$. $R^2 = 0.98$.}
		\end{subfigure}
		\begin{subfigure}[h]{0.48\textwidth}
			\centering
			\includegraphics[width=\textwidth]{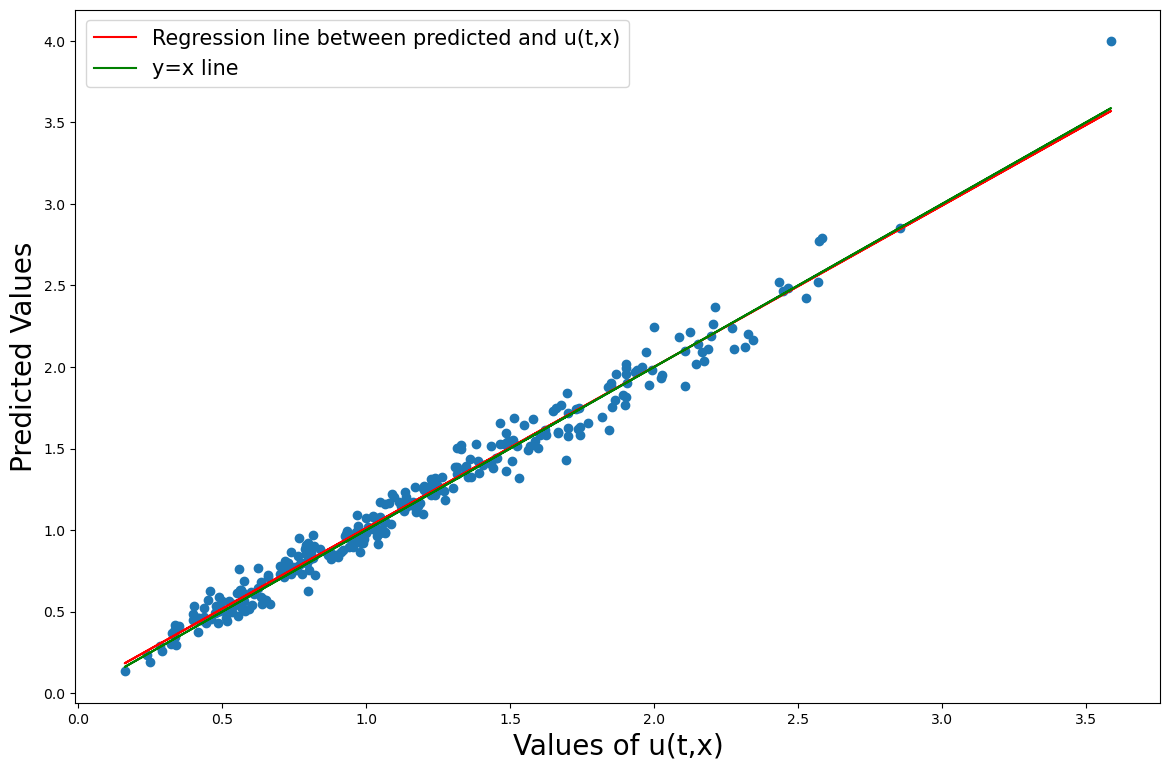}
			\caption{Prediction at $(t,x) = (1,0.5)$.\\ Relative $\ell^2$ error: $6.5\%$. Slope: $0.99$.
			Error standard deviation: $5.5\%$. $R^2 = 0.98$.}
		\end{subfigure}
\hfill
		\begin{subfigure}[h]{0.48\textwidth}
			\centering
			\includegraphics[width=\textwidth]{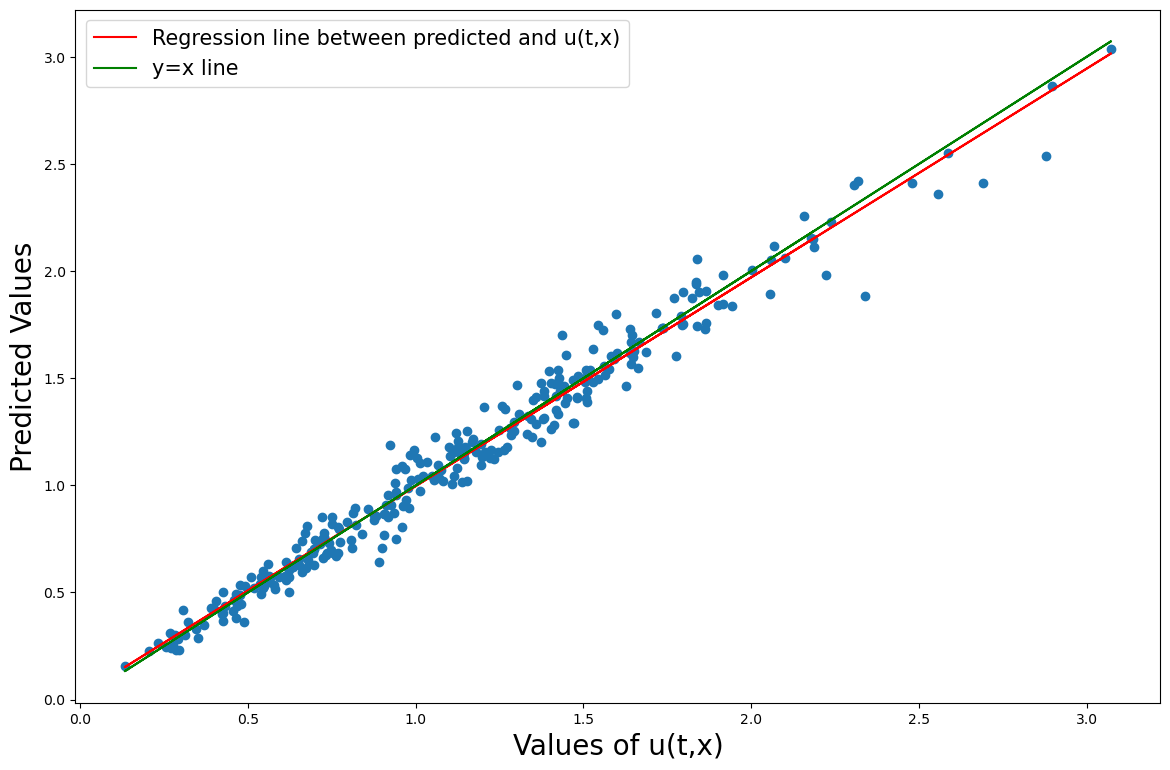}
			\caption{Prediction at $(t,x) = (1,0.95)$.\\ Relative $\ell^2$ error: $7.0\%$. Slope: $0.97$.
Error standard deviation: $6.0\%$. $R^2 = 0.97$.}
		\end{subfigure}
		\caption{Results of linear regression of solutions to~\eqref{eq:phi4m} against the functions in model $\CM^4_\alpha$ with $\alpha = (3,2,0)$, without initial conditions and of degree $\leq 5$. The $x$-axis contains values of $u(t,x)$ for realisations of the forcing $\xi$ from the test set and the $y$-axis contains predictions of the linear regression. Subplots (a), (b), (c), (d) show predictions at space-time points $(t,x) = (0.05,0.5),(0.5,0.5),(1,0.5),(1,0.95)$ respectively.}\label{fig:mult_parabolic}
	\end{figure}

	We randomly sample $1000$ realisations of approximations of white noise $\xi$ on $\CO$ and solve~\eqref{eq:phi4m} for each realisation.	
	We then split the pairs $(u,\xi)$ into training and test sets of size $700$ and $300$ respectively. There are only $56$ functions in $f_\tau \in \CM^4_\alpha$ with degree $\deg\tau \leq 5$ thus corresponding to a ratio of training cases to the number of predictors of $700/56 = 12.5$. In Figure~\ref{fig:mult_parabolic}, we show results of performing Algorithm~\ref{alg:1} with models without initial conditions at various space-time points $(t,x)$. In every subfigure one can see a scatter plot of actual values of $u(t,x)$ from the test set plotted against the predicted values.
The error is measured as a relative $\ell^2$ error,
i.e. for the vector of realisations $R$ and predictions $P$ we set
\begin{equ}
\|R-P\|_{\bar{\ell}^2} := \sqrt{\frac{1}{n}\sum_{i=1}^n |R_i-P_i|^2} \quad\text{and}\quad\|R\|_{\bar{\ell}^2} := \sqrt{\frac{1}{n}\sum_{i=1}^n |R_i|^2}\;.
\end{equ}
and the relative $\ell^2$ error is defined by
\begin{equ}
E: = Error(R,P) = \frac{\|R-P\|_{\bar{\ell}^2}}{\|R\|_{\bar{\ell}^2}} = \sqrt{\frac{\sum_{i=1}^n |R_i-P_i|^2}{\sum_{i=1}^n |R_i|^2}}\;.
\end{equ}
We also report the $R^2$ coefficient of determination and the ``error standard deviation'' which we define as
\begin{equ}
\sigma := \sqrt{\frac{1}{n}\sum_{i=1}^n \Big(E - \frac{|R_i-P_i|}{\|R\|_{\bar{\ell}^2}}\Big)^2}\;.
%= \sqrt{\frac{1}{n}\sum_{i=1}^n (\|R-P\|_{\bar{\ell}^2} - |R_i-P_i|)^2}/\|R\|_{\bar{\ell}^2}
\end{equ}
We also report the slope of the regression line between true values and the predicted ones.

	In Figure~\ref{fig:mult_parabolic} one sees a better fit for a small time $t = 0.05$ which is explained by the fact that the approximation of $u$ by functions from $\CM^4_\alpha$ is local because of the Taylor expansions in the Picard iterations (see Section~\ref{sec:motivation_SPDE} and equation \eqref{eq:double_Picard}). For larger times $t \in \{0.5,1\}$ as well as different spatial points $x \in \{0.5, 0.95\}$
there seem to be no big statistical difference in accuracy.

	\begin{table}[h]
\centering
\begin{tabular}{ |c|c|c|c|c|c|c|  }
 \hline
 & \multicolumn{3}{|c|}{$(t,x)=(0.05,0.5)$}
& \multicolumn{3}{|c|}{$(t,x)=(0.5,0.5)$}  \\
 \hline
Model height & Error & Slope & $R^2$ &
 Error & Slope & $R^2$
\\
 \hline
 1   & 9.3\%    & 0.91&  0.91 & 21.1\%    & 0.85&  0.84\\
 2&   5.4\%  & 0.97  & 0.97 & 9.5\%    & 0.97&  0.97\\
 3 & 4.9\% &  0.98 &  0.97 & 8.0\%    & 0.98&  0.98 \\
 4    & 4.8\% & 0.98 &  0.98 & 7.7\%    & 0.98&  0.98 \\
\hline
& \multicolumn{3}{|c|}{$(t,x)=(1,0.5)$}
& \multicolumn{3}{|c|}{$(t,x)=(1,0.95)$}  \\
 \hline
Model height & Error & Slope & $R^2$ &
 Error & Slope & $R^2$
\\
 \hline
 1   & 23.2\%    & 0.73 &  0.73 & 22.1\% &  0.75 &  0.75\\
 2&   13.7\% & 0.91  & 0.91 & 13.4\% &  0.91 &  0.91\\
 3 & 7.7\% & 0.97 &  0.97 & 7.7\% &  0.97 &  0.97\\
 4    & 6.5\% & 0.98 &  0.98 & 6.6\% &  0.98 &  0.98 \\
\hline
\end{tabular}
%		\centering
%		\includegraphics[width=\textwidth]{PtableColor}
		\caption{Average relative $\ell^2$ errors, slopes, and $R^2$ for linear regression against models of different heights. Prediction is performed at space-time points $(t,x) = (0.05,0.5),(0.5,0.5),(1,0.5),(1,0.95)$.}\label{table:mult_parabolic}
	\end{table}

	In Table~\ref{table:mult_parabolic}, we show average relative $\ell^2$ error, slope of the regression line, and $R^2$ statistic for Algorithm~\ref{alg:1} applied to models of heights $1,2,3$ and $4$.
All experiments are performed $1000$ times (i.e. splitting the data randomly into training/test sets) and the average values over these experiments are reported.
Table~\ref{table:mult_parabolic} demonstrates that increasing height indeed allows for a better overall prediction. A similar result holds true for the width: additive width smaller than $3$ (which corresponds to the third power in the non-linearity in~\eqref{eq:phi4m}) gives on average a worse error.
	
	\begin{remark}\label{rem:reduced_model}
		Note that the error for the middle time $t = 0.5$ is slightly worse than for the end time $t = 1$.
This could be caused by using a model without initial conditions instead of the full model. Indeed, using the full model as in Definition~\ref{def:model_heat} allows to slightly reduce the error for the prediction at $(t,x) = (0.5, 0.5)$ to $7.4\%$ (with the same $n = 4$ and $\alpha = (3,2,0)$) while making almost no change to the error for the prediction at $(t,x) = (1, 0.5)$ and at $(t,x) = (0.05, 0.5)$.

		A heuristic reason why the effect of the fixed initial condition could be ignored for the parabolic equations could be a good local structure and dissipative properties of the heat operator. The advantage of using models with $\CJ = \emptyset$ for parabolic equations is that such models contain fewer functions, which both improves the speed of the computation and potentially helps with problems of overfitting.
	\end{remark}

We compare the results from Algorithm~\ref{alg:1} with several basic off-the-shelf learning algorithms. To do this,
	 for $(u,\xi) \in U^\obs$ we transform all the space-time points of the forcing $\xi$ into a vector (in this case a vector of $100000$ points) and applied support vector regression (SVR), K-nearest neighbours (KNN), and random forest regressions (RFR) to predict the value of $u(t,x)$ for $(t,x)=(1,0.5)\in [0,T]\times D$.
These algorithms were applied with the default settings in the Python \verb|sklearn| library (e.g. SVR was taken with the RBF kernel)
and each algorithm was tested on $1000$ realisations of the noise with a $700$:$300$ split for training and testing data as before.
We give the results in Table \ref{table:off_shelf}.
None of these algorithms gave better than $35\%$ error.
We also subsampled $\xi$ by taking $50, 200$ and $1000$ evenly sampled space-time points (vs. the full $100000$ points) to avoid over-fitting, but these three choices only increased the error for each regressor.
%Moreover, neither scaling the data, nor taking less space-time points of $\xi$, nor increasing the number of functions in $U^\obs$ drastically improved the error.
This short comparison demonstrates that the MFV captures information that is lost by treating the noise simply as a large vector.

	\begin{table}[h]
\begin{tabular}{ |c|c|c|c|c|c|c|c|c| }
 \hline
\multicolumn{3}{|c|}{RFR} & \multicolumn{3}{|c|}{SVR}
& \multicolumn{3}{|c|}{KNN}  \\
 \hline
Error & Slope & $R^2$ &
 Error & Slope & $R^2$ & Error & Slope & $R^2$
\\
 \hline
42.7\%    & 0.04 & 0.03  &
37.2\%  & 0.23 &  0.30 &
44.2\%    & 0.12 & 0.003 \\
\hline
\end{tabular}
		\centering
		\caption{Average relative $\ell^2$ errors, slopes, and $R^2$ for off-the-shelf learning algorithms applied to flattened noise with $100000$ points. Prediction is performed at space-time point $(t,x) = (1,0.5)$.}\label{table:off_shelf}
	\end{table}
	
\subsubsection{Two-dimensional spatial domain}
\label{subsubsec:2D}

We consider a similar experiment as in the previous subsection but over a two-dimensional domain.
Specifically, we consider the PDE
	\begin{equs}[eq:phi4m_2D]
		(\partial_t - \Delta) u &= 3u - u^3 + u\,\xi\,\quad\text{for $(t,x) \in [0,1]\times \T^2$,}\\
		u(0,x,y) &= \cos(2\pi(x+y)) + \sin (2\pi(x+y))\,,
	\end{equs}
where $\T^2 \eqdef \R^2/\Z^2$ is the two-dimensional torus that we identify with $[0,1)^2$ as a set (i.e. we consider periodic boundary conditions).
We take the forcing $\xi$ now as white in time and coloured in space. The precise definition is $\xi(t,x,y) = \dot{\beta}(t) w(x,y)$ where $\beta(t)$ is Brownian motion, and $w$ is independent of $\beta$ and normally distributed according to $\mathcal{N}(0, 3^{3/2} (-\Delta + 49I)^{-3})$ as in \cite[Sec 4.3]{NEURIPS2022_09116662}. Here $\Delta$ is a periodic Laplacian on $\T^2$ and we take its discrete periodic approximation to generate the data.
The reason why we take a coloured noise in space instead of space-time white noise is that the equation \eqref{eq:phi4m_2D} is singular in two spatial dimensions (see \cite{Regularity}) and does not have a classical solution if the noise is white in both space and time. 

We discretise space into $64\times 64$ points and time into $1000$ points.
We construct a model with $\CJ = \{c\}$ a singleton and height $n=2$ and remaining parameters ($m,\ell,q$ and degree cut-off) as in Section \ref{sec:parabolic_multi}.
For the corresponding function $u^{c}$ in \eqref{eq:signal} we take $u^{(c)}=I_c[u_0]+I[\xi]$
where $I_c$ and $I$ are as in \eqref{eq:SHE}.
Remark that this definition slightly differs from Definition \ref{def:model_heat}; we made this choice to still incorporate the initial condition while keeping the number of functions in $\CM^n_\alpha$
relatively small for computational reasons (cf. Remark \ref{rem:reduced_model}). 
Note though that this implies that the height $0$ model already has some non-trivial information (coming from $I[\xi]$).

\begin{figure}[h]
\centering
		\begin{subfigure}[h]{0.48\textwidth}
			\centering
			\includegraphics[width=\textwidth]{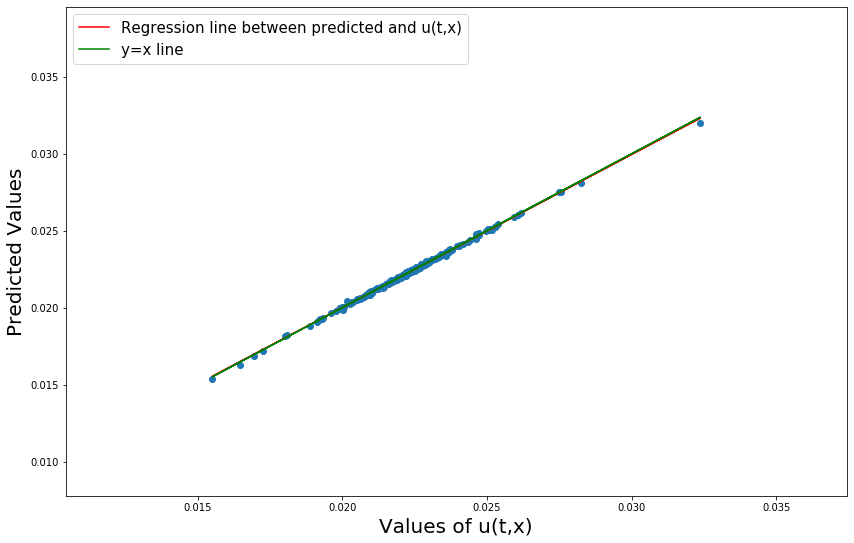}
			\caption{Prediction at $(t,x,y) = (0.05,0.5,0.5)$.\\ Relative $\ell^2$ error: $0.26\%$. Slope: $0.995$. Error standard deviation: $0.26\%$. $R^2 = 0.999$}
		\end{subfigure}
\hfill
		\begin{subfigure}[h]{0.48\textwidth}
			\centering
			\includegraphics[width=\textwidth]{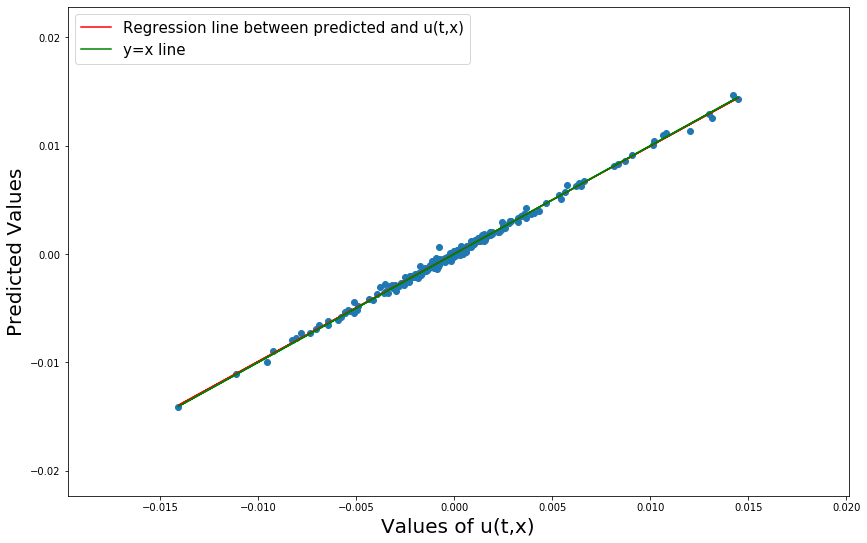}
			\caption{Prediction at $(t,x,y) = (0.5,0.5,0.5)$.\\ Relative $\ell^2$ error: $6.3\%$. Slope: $0.994$. Error standard deviation: $6.3\%$. $R^2 = 0.996$.}
		\end{subfigure}
		\begin{subfigure}[h]{0.48\textwidth}
			\centering
			\includegraphics[width=\textwidth]{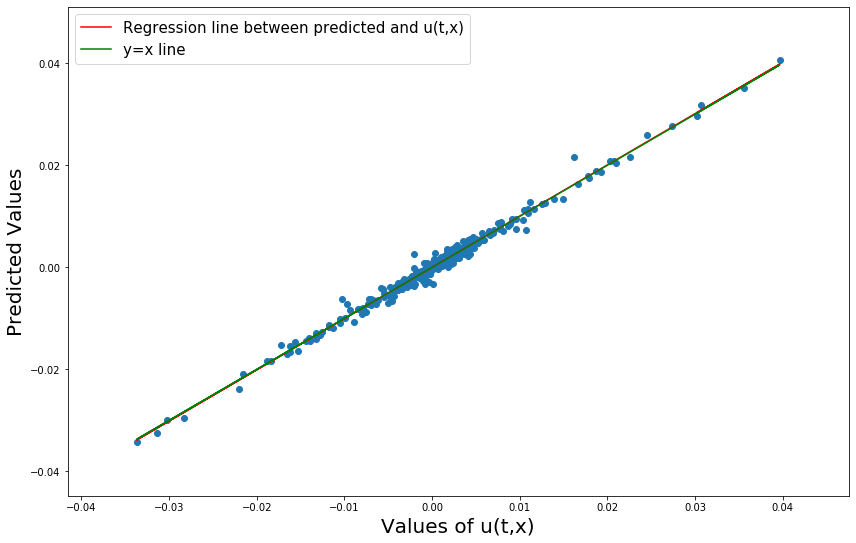}
			\caption{Prediction at $(t,x,y) = (1.0,0.5,0.5)$.\\  Relative $\ell^2$ error: $10.9\%$. Slope: $1.005$. Error standard deviation: $10.8\%$. $R^2 = 0.988$.}
		\end{subfigure}
		\caption{Results of linear regression of solution to~\eqref{eq:phi4m_2D}. The $x$-axis contains values of $u(t,x,y)$, where $(t,x,y)$ is the indicated space-time point, from the test set and the $y$-axis contains predictions of the linear regression.}\label{fig:2D}
	\end{figure}

We sampled $1000$ realisations of $\xi$ and performed Algorithm \ref{alg:1} as in Section \ref{sec:parabolic_multi} (with $700$ training and $300$ testing samples).
Figure \ref{fig:2D} shows the outcome for three space-time points.
As in Figure \ref{fig:mult_parabolic}, we see an decrease in accuracy at larger times, which is expected from theory (as explained in Section \ref{sec:parabolic_multi}).
We also mention that we performed the same experiment with no initial conditions (i.e. with $\CJ=\emptyset$ as Section \ref{sec:parabolic_multi}), but achieved no meaningful predictive power (see Section \ref{sec:Wave} for a similar outcome for the wave equation),
a feature not encountered in the one-dimensional setting of Section \ref{sec:parabolic_multi}.

We furthermore performed the experiment $1000$ times, each time resampling the training and testing set,
and record the averages of the relative $\ell^2$ error, slope of regression line,
and $R^2$ statistic.
The results are recorded in Table~\ref{table:2D}.
As in Table~\ref{table:mult_parabolic}, we see a sharp rise in predictive power with the height of the model, further demonstrating that non-linearities in the MFV capture important information of the underlying signal.
We note that height $n=2$ here effectively corresponds to height $n=3$ of Section \ref{sec:parabolic_multi} since we included $I[\xi]$ in $u^{(c)}$.

	\begin{table}[h]
\centering
\begin{tabular}{ |c|c|c|c|c|c|c|  }
 \hline
 & \multicolumn{3}{|c|}{$(t,x,y)=(0.05,0.5,0.5)$}
& \multicolumn{3}{|c|}{$(t,x,y)=(0.5,0.5,0.5)$}  \\
 \hline
Model height & Error & Slope & $R^2$ &
 Error & Slope & $R^2$
\\
 \hline
 0   & 7.0\%    & 0.06&  0.05 & 100.3\%    & 0.00&  0.00\\
 1&   1.3\%  & 0.97  & 0.97 & 46.8\%    & 0.79&  0.78\\
 2 & 0.27\% &  0.999 &  0.999 & 6.1\%    & 0.996&  0.996 \\
\hline
& \multicolumn{3}{|c|}{$(t,x)=(1,0.5,0.5)$}
  \\
\cline{1-4}
Model height & Error & Slope & $R^2$ 
\\
\cline{1-4}
 0   & 100.1\%    & 0.00 &  0.00 \\
 1&  65.2\% & 0.59  & 0.57 \\
 2 & 9.8\% & 0.991 &  0.990 \\
\cline{1-4}
\end{tabular}
%		\centering
%		\includegraphics[width=\textwidth]{PtableColor}
		\caption{Average relative $\ell^2$ errors, slopes, and $R^2$ for linear regression against models of different heights for equation \eqref{eq:phi4m_2D}. Prediction is performed at the same space-time points as in Figure~\ref{fig:2D}.}\label{table:2D}
	\end{table}

	\subsubsection{Additive forcing}
	
	We repeat the same experiment for the additive version of the equation~\eqref{eq:phi4m} namely:
	\begin{equs}[eq:phi4a]
		(\partial_t - \Delta) u &= 3u - u^3 + \xi\,\quad\text{for $(t,x) \in [0,1]\times [0,1]$,}\\
		u(t,0) &= u(t,1)\quad\text{(Periodic BC),}\\
		u(0,x) &= x(1-x)\,.
	\end{equs}
	
\begin{figure}[h]
		\centering
		\begin{subfigure}[h]{0.49\textwidth}
			\centering
			\includegraphics[width=\textwidth]{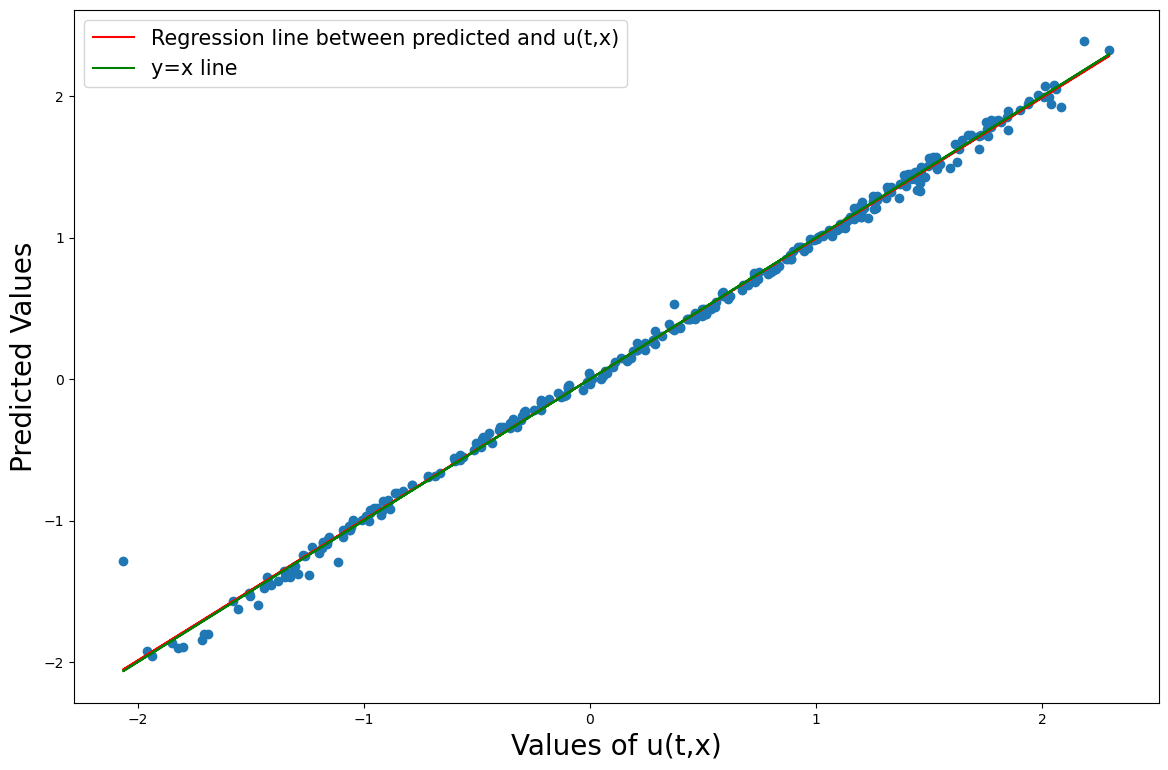}
			\caption{Prediction at $(t,x) = (0.5,0.5)$.\\ Relative $\ell^2$ error: $5.7\%$. Slope: $0.995$.
Error standard deviation: $5.6\%$. $R^2 = 0.997$.}
		\end{subfigure}
\hfill
		\begin{subfigure}[h]{0.49\textwidth}
			\centering
			\includegraphics[width=\textwidth]{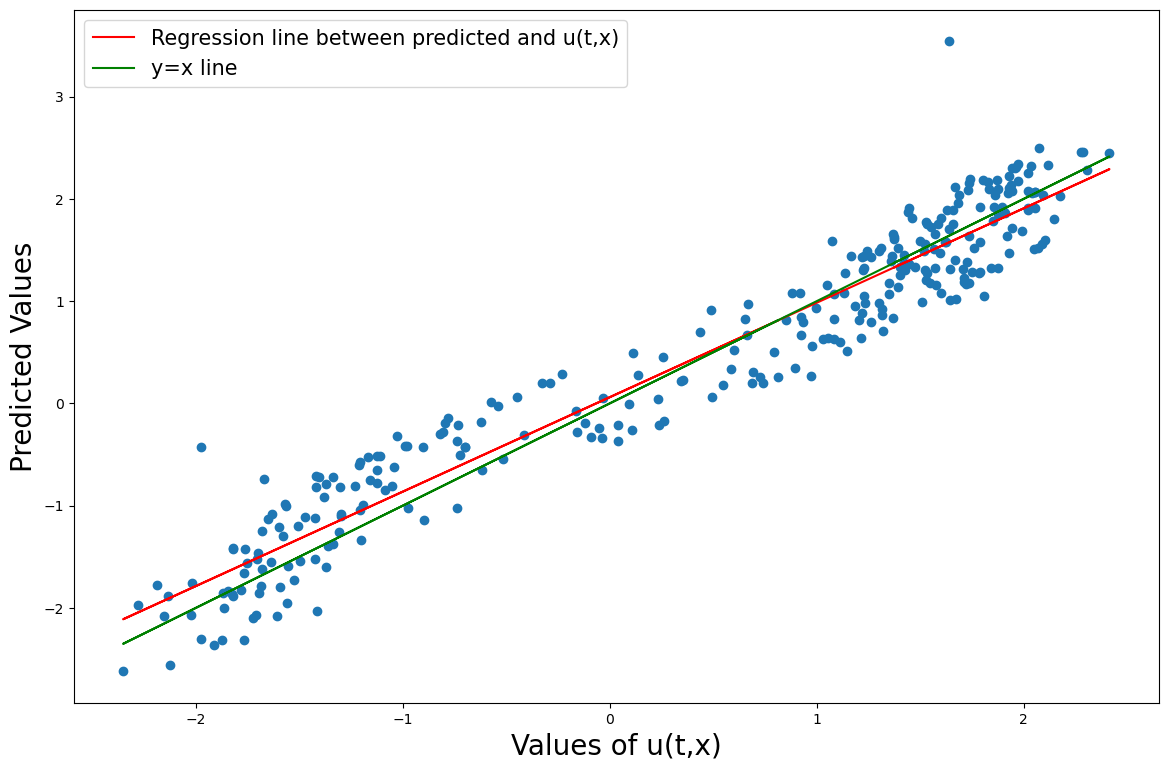}
			\caption{Prediction at $(t,x) = (1,0.5)$.\\ Relative $\ell^2$ error: $25.6\%$. Slope: $0.92$. Error standard deviation: $22.8\%$. $R^2 = 0.93$.}
		\end{subfigure}
		\caption{Results of linear regression of solution to~\eqref{eq:phi4a} against the functions in model $\CM^5_{\alpha}$ with $\alpha = (3,1,0)$, without initial conditions and of degree $\leq 7.5$. The $x$-axis contains values of $u(t,x)$ for realisations of the forcing $\xi$ from the test set and the $y$-axis contains predictions of the linear regression. Subplots (a), (b) show predictions at space-time points $(t,x) = (0.5,0.5)$ and $(t,x) = (1,0.5)$ respectively.}\label{fig:3}
	\end{figure}

	Discretisation of space-time and number of training and test cases is the same as in Section~\ref{sec:parabolic_multi}.
	We perform Algorithm~\ref{alg:1} using the model $\CM^5_{\alpha}$ from Definition~\ref{def:model_heat} with viscosity $\nu=1$ and $\alpha = (3,1,0)$, without initial conditions ($\CJ = \emptyset$), and with degree $\leq 7.5$, which gives $58$ functions.\footnote{See Remark~\ref{rem:polinoms} for a motivations behind taking these particular widths.} Note that since multiplicative width is $1$ this reduces the number of functions in the model compared to the multiplicative case of Section \ref{sec:parabolic_multi}, which allows us to take a larger height and upper bound for the degree. Figure~\ref{fig:3} shows the results for space-time points $(t,x)  = (0.5,0.5)$ and $(t,x)  = (1,0.5)$. One can see that the additive equation exhibits a worse prediction for long times compared to the multiplicative equation (Figure \ref{fig:mult_parabolic} and Table \ref{table:mult_parabolic}) but a slightly better prediction for short times (for $t = 0.05$ error is even better: $0.1\%$ in comparison to $\approx 5\%$ in the multiplicative case). 
	
The prediction error rises in this additive case as $t$ increases.
This is expected by a similar reason as mentioned in Section \ref{sec:parabolic_multi}, which is the Taylor expansions in the Picard iterations approximating $u$ (see \eqref{eq:double_Picard}).
It would be of interest to extend Algorithm \ref{alg:1}
to decrease this error.
A potential way to do this is to generalise and combine Algorithms \ref{alg:1} and \ref{alg:2} and compute models on subintervals to
learn the `flow' of the equation,
i.e. build models from the initial condition and the forcing over subintervals of $[0,T]$, use this model to predict the solution over subintervals, and chain the predictions together.
(See also the discussion at the end of Section \ref{subsubsec:predict_one_point}.)
%This would mimic the experiment in Section \ref{sec:numerics_Burgers} below.
We leave this generalisation for a future work.

	\subsection{Wave equation with forcing}\label{sec:Wave}
	
	We will now consider a wave equation taking $\mathcal{L} = \partial^2_t - \Delta$ in~\eqref{eq:PDE} and predict solutions of the following non-linear wave equation
	\begin{equs}[eq:wave]
		(\partial^2_t - \Delta) u &= \cos(\pi\, u) + u^2 + u\,\xi\,\quad\text{for $(t,x) \in [0,1]\times [0,1]$,}\\
		u(t,0) &= u(t,1)\quad\text{(Periodic BC),}\\
		u(0,x) &= \sin(2\pi\,x),\\
		\partial_t u(0,x) &= x(1-x)\,,
	\end{equs}
	where $\xi$ is a space-time forcing which we again take as a realisation of white noise. We will compare Algorithm~\ref{alg:1} with both models with and without initial conditions. Discretisation of space-time and number of training and test cases is the same as in Section~\ref{sec:parabolic_multi}.
	
	Note that, in the general case, the level zero of the full model ${\CM}^0_\alpha$  for the wave equation should not only include the contribution of the initial condition $u_0$ but also the contribution of the initial speed $\partial_t u(0,x) = v_0$. This leads to the following definition.
	\begin{definition}\label{def:model_wave}
		Consider an initial condition $u_0 : D \to \R$, an initial speed $v_0 \colon D \to \R$, and a forcing $\xi \colon [0,T] \times D \to \R$, as well as $n,m,\ell \geq 1$, $q\leq 1$ and $\nu > 0$. Let $\alpha = (m,\ell, q)$.
		The model $\CM^n_\alpha$ for the wave equation with propagation speed $\sqrt{\nu}$ is constructed by taking $\CJ = \{c,s\}$ with $u^{(c)} = I_c[u_0],$ $u^{(s)} = I_s[v_0]$ where
		\begin{equs}
			\begin{cases}
				(\partial^2_t - \nu \Delta) I_c[u_0] & = \;\,0\\
				I_c[u_0](0,x) &= u_0(x)\;,\\
				\partial_t I_c[u_0](0,x) & = 0 \;.
			\end{cases}
			\qquad
			\begin{cases}
				(\partial^2_t - \nu \Delta) I_s[v_0] & = \;\,0\\
				I_s[v_0](0,x) &= 0\;,\\
				\partial_t I_s[v_0](0,x) & = v_0(x) \;.
			\end{cases}
		\end{equs}
	 Moreover, for functions $f \colon[0,T]\times D \to \R$ the operator $I[f]$ is defined to be the solution to a wave equation 
	 \begin{equ}
	 	(\partial^2_t - \nu \Delta) I[f] = f\,,
	 \end{equ}
	 with $I[f](0,x) = \partial_t I[f](0,x) = 0$.  
	\end{definition}
	Boundary conditions for the above equation are taken to be the same as boundary conditions for the underlying wave equation (in this case periodic).
	
	For this experiment we choose $n = 4$, $\alpha = (m,\ell, q) = (2,2, 0)$, and $\nu=1$, and impose the degree to satisfy $\deg \xi = -1.5$, $\deg u_0 = \deg v_0 = -0.5$ and $\beta = 1.5$ in~\eqref{eq:degree}. We choose only functions of degree $\leq 1.5$ which gives $60$ functions in $\CM^4_\alpha$. 
	\begin{figure}[h]
		\centering
		\begin{subfigure}[h]{0.49\textwidth}
			\centering
			\includegraphics[width=\textwidth]{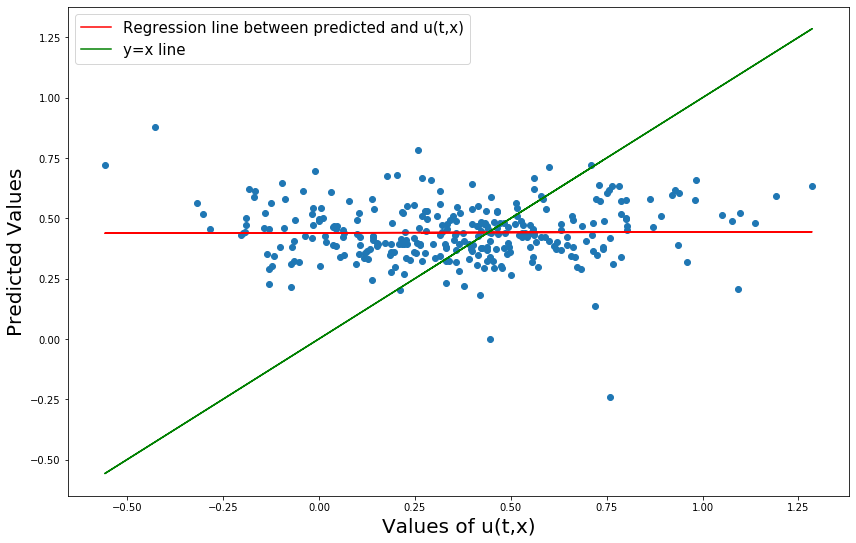}
			\caption{Prediction at $(t,x) = (1,0.5)$ for model with $\CJ = \emptyset$. Relative $\ell^2$ error: $72.1\%$.
Slope: $0.003$.
Error standard deviation: $50.7\%$. $R^2 = -0.21$.}
		\end{subfigure}
\hfill
		\begin{subfigure}[h]{0.49\textwidth}
			\centering
			\includegraphics[width=\textwidth]{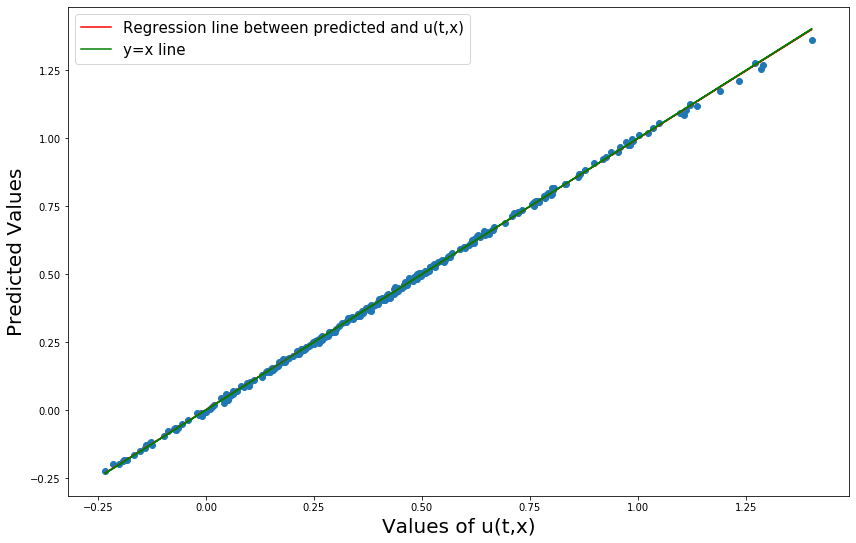}
			\caption{Prediction at $(t,x) = (1,0.5)$ for model with $\CJ = \{c,s\}$. Relative $\ell^2$ error: $1.3\%$.
Slope: $0.998$.
Error standard deviation: $0.9\%$. $R^2 = 0.999$.}
		\end{subfigure}
		\caption{Results of linear regression of solution to~\eqref{eq:wave} using functions from models $\CM^4_\alpha$ with $\alpha = (2,2, 0)$ of degree $\leq 1.5$ without and with initial conditions. The $x$-axis contains values of $u(t,x)$ for realisations of the forcing $\xi$ from the test set and the $y$-axis contains predictions of the linear regression. Subplots (a), (b) show predictions at space-time point $(t,x) = (1,0.5)$ for models with $\CJ = \emptyset$ and $\CJ = \{c,s\}$ respectively.}\label{fig:4}
	\end{figure}
	Figure~\ref{fig:4} shows the importance of using the full model in the case of the wave equation, i.e. the contribution of the initial condition (even fixed and deterministic) can't be ignored.\footnote{This parallels the necessity of including the initial condition in an analogue of the model in~\cite{singular_wave} where the authors solve a non-linear singular stochastic wave equation in $3$ dimensions.} The model constructed with $\CJ = \emptyset$ in the case of the wave equation gives absolutely no predictability contrary to the parabolic case (see Remark~\ref{rem:reduced_model}) because the wave operator is not dissipative contrary to the heat operator.

Average relative $\ell^2$ errors corresponding to different heights of the model with and without initial speed are presented in the Table~\ref{fig:wave_table} for $1000$ repeated experiments.
	Table~\ref{fig:wave_table} further shows the importance of including the contributions of both the initial condition and the initial speed in the model when predicting the wave equation.

	\begin{table}[h]
		\centering
\begin{tabular}{ |c|c|c|c|c|c|c|  }
 \hline
 & \multicolumn{3}{|c|}{With initial speed}
& \multicolumn{3}{|c|}{Without initial speed}  \\
 \hline
Model height & Error & Slope & $R^2$ &
 Error & Slope & $R^2$
\\
 \hline
 1   & 59.8\%    & 0.04&  0.03 & 59.8\%    & 0.03&  0.03\\
 2&   12.8\%  & 0.96  & 0.96 & 13.8\%    & 0.95&  0.95\\
 3 & 2.1\% &  0.999 &  0.999 & 5.0\%    & 0.994&  0.993 \\
 4    & 1.4\% & 0.999 &  0.999 & 4.3\%    & 0.995&  0.995 \\
\hline
\end{tabular}
		\caption{Average relative $\ell^2$ errors, slopes, and $R^2$ for prediction at space-time point $(t,x) = (1,0.5)$ for different heights of the model. First column involves models using $\CJ = \{c,s\}$ and the second column involves models using $\CJ = \{c\}$ only.}\label{fig:wave_table}
	\end{table}

	\subsection{Burgers' equation}\label{sec:numerics_Burgers}
	
	In this subsection, we aim to predict solutions to the following Burgers' equation with no forcing
	\begin{equs}\label{eq:Burgers}
		(\partial_t- 0.2 \Delta ) u &= - u \partial_x u\,,\quad(t,x)\in[0,10]\times [-8,8]\\
		u(t,-8) &= u(t,8)\quad\text{(Periodic BC)}\,,\\
		u(0,x) &= \sum_{k = -10}^{10} \frac{a_k}{1+|k|^2} \sin\big(\lambda^{-1} \pi k x\big)
	\end{equs}
from the knowledge of the initial condition $u_0$.
That is, given only the initial condition $u_0\colon [-8,8]\to \R$, our goal is to reconstruction the entire function $u\colon [0,10]\times[-8.8]\to \R$ without explicitly solving the PDE~\eqref{eq:Burgers}, i.e., we wish to learn the map $u_0\mapsto u$.
This experiment is partly inspired by~\cite{Deep_Burgers}.
The above equation satisfies the time homogeneous flow property that motivates Algorithm~\ref{alg:2}, which we use in the experiments below.

Above, $(a_k)_{k = -10,\dots,10}$ are sampled as independent and identically distributed (i.i.d.) standard normal random variables and $\lambda$ is a scaling parameter. We sample $120$ such initial conditions with scaling $\lambda = 8,4,2$ ($40$ initial conditions for each scaling), which corresponds to $u_0$ having respectively one, two and four cycles. We then randomly subdivide these initial conditions into training and test sets of sizes $100$ and $20$ respectively.

To discretise time, we take $201$ evenly spaced points $\CO_T\subset[0,10]$ ($N = 200$, $\delta = 0.05$ in the notation of Section~\ref{subsubsec:alg2}).
To approximate the spatial domain, we take $512$ evenly spaced points $\CO_X \subset D = [-8,8]$.
Solutions to the equation, however, we generated using a finer grid ($2001$ time points and $512$ space points).

There is no forcing $\xi$ in the equation so $\ell = 0$.
We construct the models using Definition~\ref{def:model_heat} with viscosity $\nu=0.2$.
In constructing the model, we discretise time with the finer grid, so
the relevant space-time domain in continuum is $[0,\delta]\times D$ which we discretise to $\{0,\delta/10\ldots, 9\delta/10,\delta\}\times \CO_X$.
In particular, the operator $I_c$ takes as input a function $u_0\colon\CO_X\to\R$ and outputs a function $I_c[u_0]\colon \{0,\delta/10\ldots, 9\delta/10,\delta\}\times \CO_X\to\R$.

We choose height $n = 3$, additive width $m= 2$, and differentiation order $q=1$ for the model, i.e. $\alpha = (2,0,1)$. We assign for the degree $\deg u_0 = -1.5$ and $\beta = 2$ in~\eqref{eq:degree} and only functions $f_\tau$ with degree $\deg\tau \leq 2.5$ are considered. This gives $20$ functions of $\deg\tau \leq 2.5$ instead of the original $91$ functions in $\CM^3_\alpha$.
Using this degree cutoff speeds up the fitting of the linear regression by around $20$ times in addition to a faster computation of the model.

As the number of training and testing cases ($100$ and $20$) is relatively small, we repeated the above experiment $10$ times.
In the first row of Table \ref{table:Burgers_Alg2} we record the performance of Algorithm \ref{alg:2}, namely the averages, ranges, and standard deviations of the relative $\ell^2$ error over the $10$ experiments with $20$ test cases each.
Here the relative $\ell^2$ error for one test case with true solution $R$ and predicted solution $P$ is defined as
\begin{equ}
E = \frac{\|R-P\|_{\bar \ell^2}}{\|R\|_{\bar \ell^2}}\;,\qquad
\|R\|_{\bar \ell^2}^2:=\frac{1}{201\times512}\sum_{(t,x)} |R(t,x)|^2
\end{equ}
(the sum is over the observed grid points $\CO_T\times\CO_X \subset [0,10]\times[-8,8]$).

	\begin{table}[h]
\centering
\begin{tabular}{ |c|c|c|c|c|c|c| }
 \hline
&  \multicolumn{5}{|c|}{Algorithm \ref{alg:2}} \\
\hline
 & AE & AER & TER & ASD & SDR
\\
 \hline
no noise  & 1.1\% & 0.8\%--1.6\%   & 0.06\%--9.4\%&  1.1\% & 0.8\%--1.9\% \\
 EV &   1.1\% & 0.8\%--1.6\% & 0.06\%--9.2\% & 1.1\% & 0.7\%--1.9\%\\
 1\% noise & 8.8\% & 5.8\%--12.4\% & 0.4\%--50.0\% & 8.0\% & 4.9\%--11.1\% \\
3\% noise & 13.7\% & 8.7\%--13.7\% & 1.8\%--68.3\% & 8.7\% & 4.8\%--12.8\% \\
\hline
&  \multicolumn{5}{|c|}{PDE-FIND} \\
\hline
no noise & 0.9\% & 0.5\%--2.3\%   & 0.2\%--10.5\%&  0.6\% & 0.2\%--2.2\% \\
1\% noise & 7.3\% & 4.6\%--14.2\%   & 0.9\%--32.7\%&  5.5\% & 3.3\%--8.9\% \\
3\% noise & 19.1\% & 13.9\%--25.1\%   & 3.2\%--77.1\%&  12.5\% & 7.8\%--18.4\% \\
\hline
\end{tabular}
%		\centering
%		\includegraphics[width=\textwidth]{PtableColor}
		\caption{Performance of Algorithm \ref{alg:2} on predicting solutions to \eqref{eq:Burgers}
with no noise and true viscosity $\nu=0.2$,
with estimated viscosity (EV), and with 1\% noise and 3\% noise on observed data.
Comparison is given with PDE-FIND on observations with no noise, 1\%, and 3\% noise (see Section \ref{subsubsec:comparison}).\\
AE: average relative $\ell^2$ errors over all experiments and test cases.\\
AER: average error range over all experiments.\\
TER: total error range over all experiments and test cases.\\
ASD: average standard deviation over all experiments.\\
SDR: standard deviation range over all experiments. }\label{table:Burgers_Alg2}
	\end{table}

	\begin{table}[h]
\centering
\begin{tabular}{ |c|c|c| }
 \hline
Average & Range & Standard deviation 
\\
 \hline
0.177  & 0.171--0.182 & 0.003  \\
\hline
\end{tabular}
%		\centering
%		\includegraphics[width=\textwidth]{PtableColor}
		\caption{Estimated viscosity $\tilde\nu$ via linear regression over $10$ experiments. }\label{table:Burgers_visc}
	\end{table}

\begin{figure}[h]
		\begin{subfigure}[h]{0.95\textwidth}
			\centering
			\includegraphics[width=\textwidth]{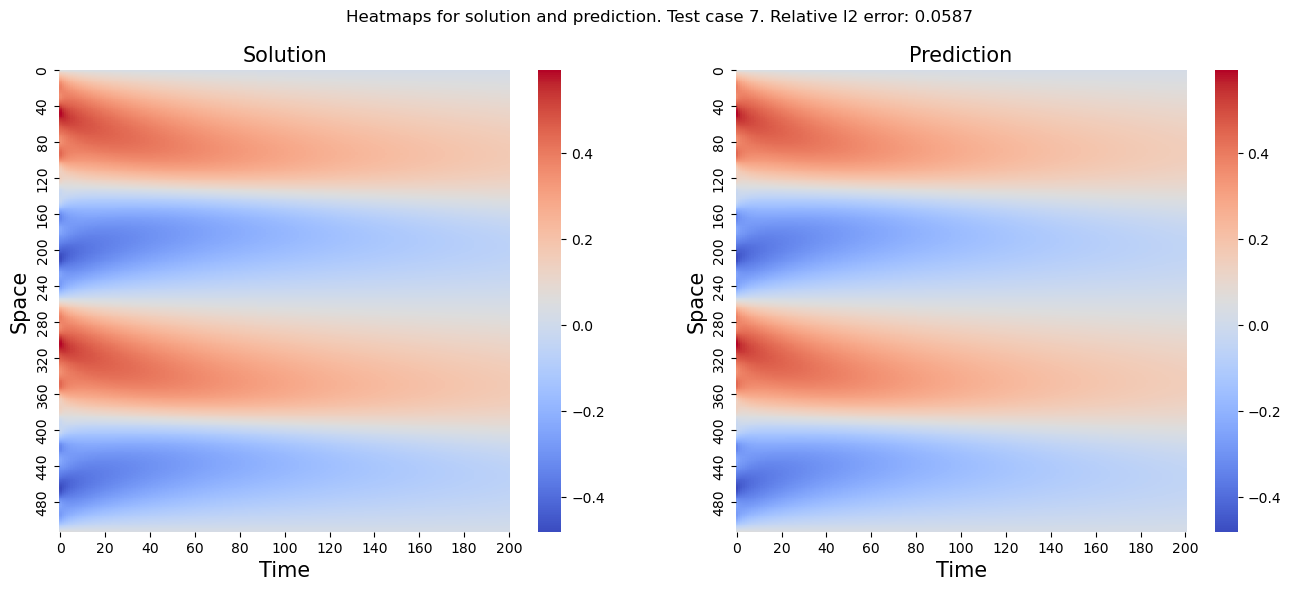}
			\caption{Relative $\ell^2$ error: $5.9\%$.}
		\end{subfigure}
%\hfill
%		\begin{subfigure}[h]{0.48\textwidth}
%			\centering
%			\includegraphics[width=\textwidth]{BEr1e4.png}
%			\caption{Relative $\ell^2$ error: $1.4\%$.}
%		\end{subfigure}
		\hfill
		\begin{subfigure}[h]{0.95\textwidth}
			\centering
			\includegraphics[width=\textwidth]{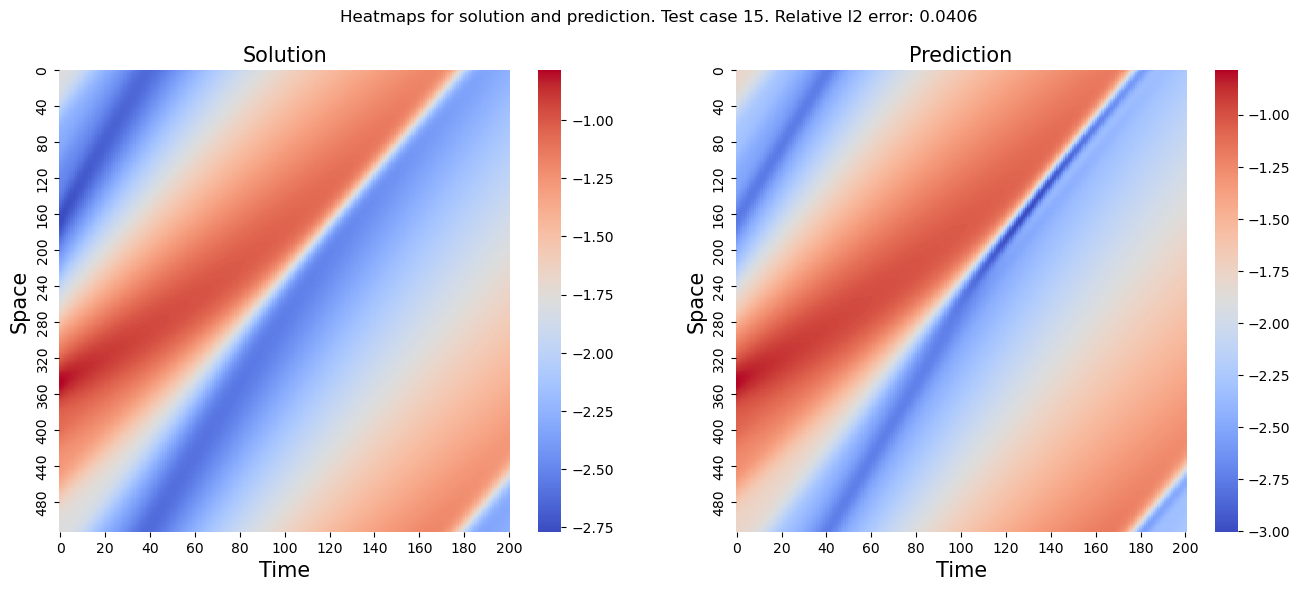}
			\caption{Relative $\ell^2$ error: $4.1\%$.}
		\end{subfigure}
%\hfill
%		\begin{subfigure}[h]{0.48\textwidth}
%			\centering
%			\includegraphics[width=\textwidth]{BEr7e9.png}
%			\caption{Relative $\ell^2$ error: $7.9\%$.}
%		\end{subfigure}
		\caption{Heat-maps for the solutions of~\eqref{eq:Burgers} (left) and predictions for two test cases (right) using Algorithm~\ref{alg:2} with functions from the model $\CM^3_\alpha$ with $\alpha = (2,0,1)$ of degree $\leq 2.5$. The values of $\lambda$ are $4$ for subfigure (a) and $8$ for (b).}\label{fig:Burgers_Alg_2_heat}
	\end{figure}

Figure~\ref{fig:Burgers_Alg_2_heat} shows the heat-maps for the true and predicted solutions drawn from two test cases with greater than average error (heat-maps for test cases with error close to the average error appeared indistinguishable to the naked eye; even for Figure \ref{fig:Burgers_Alg_2_heat}(a), where the error of $5.9\%$ is above the average, the two solutions appear similar).
	
\begin{figure}[h]
		\begin{subfigure}[h]{0.95\textwidth}
			\centering
			\includegraphics[width=\textwidth]{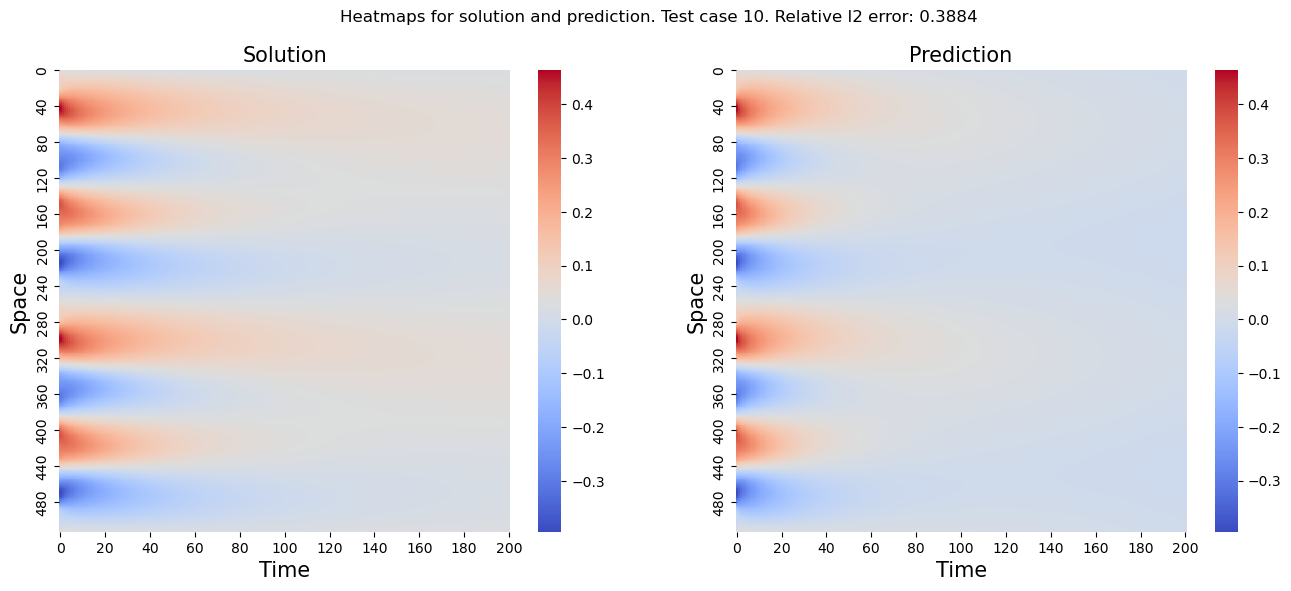}
			\caption{Relative $\ell^2$ error: $38.8\%$.}
		\end{subfigure}
%\hfill
%		\begin{subfigure}[h]{0.48\textwidth}
%			\centering
%			\includegraphics[width=\textwidth]{BEr1e4.png}
%			\caption{Relative $\ell^2$ error: $1.4\%$.}
%		\end{subfigure}
		\hfill
		\begin{subfigure}[h]{0.95\textwidth}
			\centering
			\includegraphics[width=\textwidth]{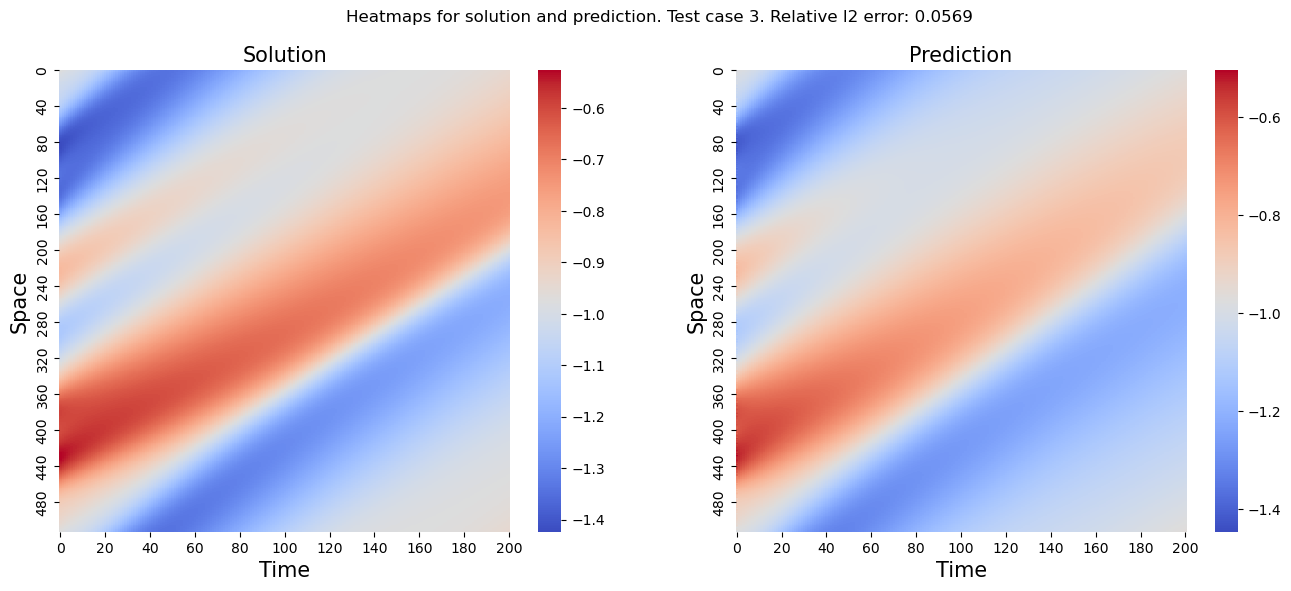}
			\caption{Relative $\ell^2$ error: $5.7\%$.}
		\end{subfigure}
%\hfill
%		\begin{subfigure}[h]{0.48\textwidth}
%			\centering
%			\includegraphics[width=\textwidth]{BEr7e9.png}
%			\caption{Relative $\ell^2$ error: $7.9\%$.}
%		\end{subfigure}
		\caption{Heat-maps for the solutions of~\eqref{eq:Burgers} (left) and predictions for two test cases (right) using Algorithm~\ref{alg:2} with $1\%$ error in observed samples. The values of $\lambda$ are $2$ for subfigure (a) and $8$ for (b).}\label{fig:Burgers_heat_noise_001}
	\end{figure}

We furthermore tested Algorithm~\ref{alg:2} on noisy data. We added a $1\%$ error (resp. $3\%$) to the observed data in the following way. Instead of observing the solution $u$ of~\eqref{eq:Burgers}, we observe
	\begin{equ}[eq:noisy]
		\tilde u(t,x) = u(t,x) + \eps(t,x)\| u\|_{\bar l^1}\;,
	\end{equ}
	where $\| u\|_{\bar \ell^1}=\frac{1}{201\times512}\sum_{(t,x)} |u(t,x)|$ (the sum is over the observed grid points $\CO_T\times\CO_X \subset [0,10]\times[-8,8]$) and $\eps(t,x)$ are i.i.d. normal random variables with zero mean and standard deviation $0.01$ (resp. $0.03$) for each $(t,x)$ for the training data and for $(t,x) = (0,x)$ for the test data.
The corresponding errors, over $10$ experiments, are presented in Table \ref{table:Burgers_Alg2}.
In Figure \ref{fig:Burgers_heat_noise_001} we give two examples of heat-maps with varying relative $\ell^2$ error for $1\%$ noise and in Figure~\ref{fig:Burgers_heat_noise_003}(a) we give an example with $3\%$ noise.
%Compared to Figure \ref{fig:Burgers_Alg_2_heat},
%we see more noticeable differences between true and predicted solutions as expected, however qualitative properties remain similar.

	Finally, we ran Algorithm~\ref{alg:2} using models in which the viscosity parameter $\nu$ in Definition~\ref{def:model_heat} is estimated from the data (the true value being $\nu=0.2$).
	We estimate $\nu$ by simply linearly regressing the discrete time derivative of $u$ against the discrete second derivative in space for $u \in U^\obs$. To be more precise, we use ordinary least squares to determine the best $\tilde\nu$ that fits
	\begin{equ}
		(u_{t_1}(x) - u_{t_0}(x))/\delta = \epsilon + \tilde{\nu} \partial^2_x u_{t_0}(x)\,,\quad u \in U^\obs\,,\;x\in D\,,
	\end{equ}
	where $\partial^2_x$ is computed using central finite difference. Such an estimate does not require any knowledge of the non-linearity.
(One could further use cross-validation to find a better estimate for $\nu$ from the interval  $[\tilde{\nu} - a, \tilde{\nu} + a]$ for some $a > 0$, although we did not do this.)
The average, range, and standard deviation of the estimated viscosity over the $10$ experiments
is recorded in Table~\ref{table:Burgers_visc}.
	
We record the errors in Table \ref{table:Burgers_Alg2} for Algorithm \ref{alg:2} with these estimated viscosities.
We find that the results are essentially the same as those with the correct viscosity $\nu = 0.2$ (and in fact have tiny improvements over the latter).

	\subsubsection{Comparison with PDE-FIND algorithm}
	\label{subsubsec:comparison}

	We use a version of PDE-FIND algorithm from~\cite{PDE-FIND} to learn the non-linearity first instead of learning the solution. We use linear regression to find the best coefficients $a,b$ such that
	\begin{equ}[eq:PDE-FIND]
		\partial_t u(t_k,x) = a \partial^2_x u(t_k,x) + b u(t_k,x)\partial_x u(t_k,x)\;,
	\end{equ}
for $u \in U^\obs$, $k = 0,\dots N-1$, and $x \in D$,
	where $\partial_t u(t_k,x) := \frac{u(t_{k+1},x)-u(t_{k},x)}{\delta}$ is a discrete time derivative, $\partial_x$ is discrete space derivative and $x \in D$ are the observed points.

	\begin{table}[h]
\centering
\begin{tabular}{|c|c|c|c|c|c|c| }
 \hline
&  \multicolumn{3}{|c|}{$a$ }  & \multicolumn{3}{|c|}{$-b$} \\
\hline
& Average & Range & SD & Average & Range & SD 
\\
 \hline
no noise & 0.212  & 0.203--0.246 & 0.012 & 0.984 & 0.979--0.987 & 0.002 \\
1\% noise & 0.115 & 0.085--0.139 & 0.016 & 0.902 & 0.742--0.952 & 0.058 \\
3\% noise & 0.027 & 0.021--0.038 & 0.005 & 0.733 & 0.640--0.796 & 0.045 \\
\hline
\end{tabular}
%		\centering
%		\includegraphics[width=\textwidth]{PtableColor}
		\caption{Estimated parameters $a$ and $-b$ in \eqref{eq:PDE-FIND} via PDE-FIND. The true values are $a=0.2$ and $-b=1$.}\label{table:PDE_FIND_coef}
	\end{table}

We then use finite difference method with the estimated $(a,b)$ (estimating these coefficients separately for each experiment) starting from initial conditions from $U^\pr$ in order to construct the predicted solution on the full domain $[0,10]\times D$.
Note though that in this finite difference we can discretise time on a finer grid. In fact, we take $2001$ points on $[0,10]$ which is the same number of time points that is used to construct solution to~\eqref{eq:Burgers}.
	We cannot, however, discretise the spatial domain $[-8,8]$ to a finer grid than $512$ points because these are the only points observed for the initial conditions from $U^\pr$.

The estimated coefficients $a,b$ are recorded in Table~\ref{table:PDE_FIND_coef}.
The resulting errors for the predicted solutions after repeating the experiment $10$ times with $20$ test cases as before are recorded in Table \ref{table:Burgers_Alg2}.
We notice that the performance of Algorithm \ref{alg:2} (with and without estimated viscosity)
is similar to that of PDE-FIND, although Algorithm \ref{alg:2} yields a slightly larger average error, but a slightly lower maximum average error and total error.

We furthermore performed the same experiments but with 1\% noise and 3\% noise on observed samples as for Algorithm \ref{alg:2}.
Here, instead of direct linear regression,
we follow the proposal in \cite{PDE-FIND} and perform polynomial interpolation: for each space-time point $z$, we fit a polynomial of degree $4$ that best matches the observed function in a neighbourhood of radius $20$ points around $z$,
and then estimate $a,b$ in \eqref{eq:PDE-FIND} by taking derivatives of these polynomials and applying linear regression.
This is made in order to avoid taking explicit derivatives of $\tilde u$ via the finite difference method since the noisy data is not differentiable.
The resulting errors and estimates for $a,b$ are recorded in Tables \ref{table:Burgers_Alg2} and \ref{table:PDE_FIND_coef} respectively.
On 1\% noise, the two methods are again comparable (with PDE-FIND demonstrating a slightly lower average error).

\begin{figure}[h]
		\begin{subfigure}[h]{0.95\textwidth}
			\centering
			\includegraphics[width=\textwidth]{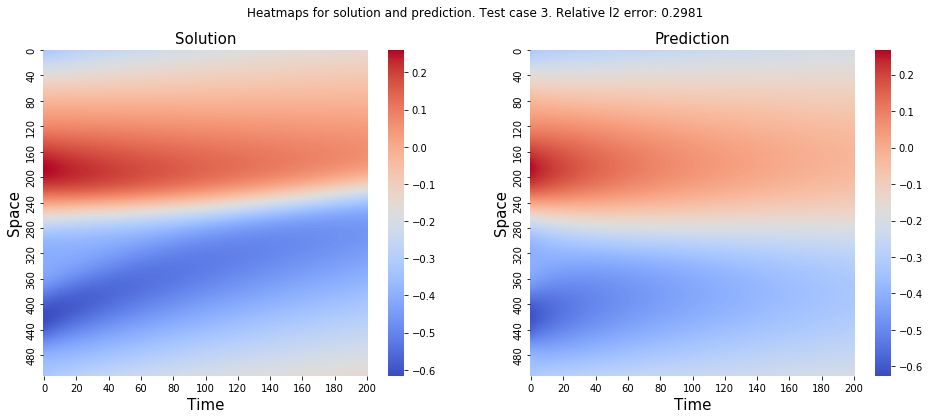}
			\caption{Algorithm \ref{alg:2}, 3\% noise, relative $\ell^2$ error: $29.8\%$.}
		\end{subfigure}
%\hfill
%		\begin{subfigure}[h]{0.48\textwidth}
%			\centering
%			\includegraphics[width=\textwidth]{BEr1e4.png}
%			\caption{Relative $\ell^2$ error: $1.4\%$.}
%		\end{subfigure}
		\hfill
		\begin{subfigure}[h]{0.95\textwidth}
			\centering
			\includegraphics[width=\textwidth]{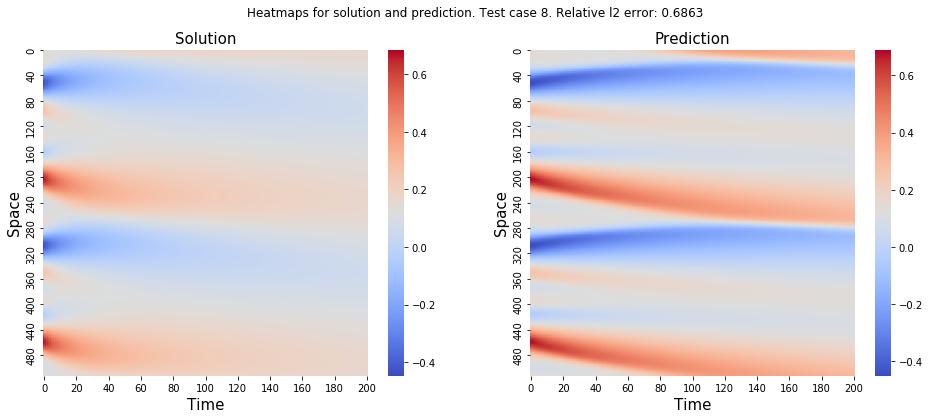}
			\caption{PDE-FIND, 3\% noise, relative $\ell^2$ error: $68.6\%$.}
		\end{subfigure}
%\hfill
%		\begin{subfigure}[h]{0.48\textwidth}
%			\centering
%			\includegraphics[width=\textwidth]{BEr7e9.png}
%			\caption{Relative $\ell^2$ error: $7.9\%$.}
%		\end{subfigure}
		\caption{Heat-maps for the solutions of~\eqref{eq:Burgers} (left) and predictions for two test cases (right) with 3\% noise on observed data. Subfigure (a) is for Algorithm~\ref{alg:2} and subfigure (b) is for PDE-FIND. The values of $\lambda$ are $8$ for subfigure (a) and $4$ for (b).}\label{fig:Burgers_heat_noise_003}
	\end{figure}

However, with 3\% noise, we found that there is a noticeable difference. First, the estimated viscosity $a$ 
in Table \ref{table:PDE_FIND_coef} is between $0.02$ and $0.04$, which is significantly lower than the true value $0.2$.
This caused the predicted solution to blow up on some test cases (due to numerical instability in our finite difference method): in each of the $10$ experiments, between $0$ and $7$ of the $20$ test cases blew up (the two extreme values were attained only for one experiment each, and the most common number of blow-ups was $1$). 
%2 0 1 4 1 3 7 1 1 2
Figure~\ref{fig:Burgers_heat_noise_003}(b) shows heat-maps for a non-blow-up test case with 3\% noise using PDE-FIND,
wherein one can see the effect of the low estimated viscosity.
In comparison, no test cases for Algorithm \ref{alg:2} blew up.

In Table \ref{table:Burgers_Alg2} we only report errors from test cases where PDE-FIND did \textit{not} blow up -- the errors are expected to be even larger if all test cases were included by solving the associated equation with a more sophisticated numerical scheme.
Even after removing the test cases for which PDE-FIND blew up, we see a mild advantage of Algorithm \ref{alg:2} over PDE-FIND with polynomial interpolation.

Finally, the reader may wonder if it is fair to compare Algorithm \ref{alg:2} to PDE-FIND given that we input into Algorithm \ref{alg:2} the true viscosity $0.2$, while PDE-FIND is required to estimate it.
We point out, however, that Algorithm \ref{alg:2} has no knowledge of the non-linearity $u\partial_x u$ in \eqref{eq:Burgers} (though the parameters $m,q$ are chosen with the motivation that the non-linearity is at most quadratic with $\partial_x u$ possibly appearing),
while in our implementation of PDE-FIND we do input $u\partial_x u$ as the only possible non-linearity.
Furthermore, on noiseless data, the viscosity estimated from the data gives results for Algorithm \ref{alg:2} that are comparable to PDE-FIND (see Table \ref{table:Burgers_Alg2}).
We also point out that Algorithm \ref{alg:2} was approximatively $20$ times faster to run with a fast Fourier transform method of computing models than
PDE-FIND with polynomial interpolation.

	\section{Summary and discussion}
	\label{sec:discussion}

To summarise, we proposed a new \textit{model feature vector} (MFV) of a space-time signal that extends to multi-dimensional space the notion of a path signature.
We further proposed two regression algorithms, which reveal that MFVs may contain important information about the underlying signal.

We applied Algorithm~\ref{alg:1} to both parabolic and hyperbolic equations with forcing and Algorithm~\ref{alg:2} to Burgers' equation with varying initial conditions.
We did an elementary comparison of our algorithms with other methods.
We compared the performance of Algorithm~\ref{alg:1} for the parabolic equation with multiplicative forcing against several off-the-shelf methods and found a large advantage in favour of Algorithm~\ref{alg:1}.
We further compared Algorithm~\ref{alg:2} for the Burgers' equation against a version of PDE-FIND~\cite{PDE-FIND} (see Section~\ref{subsubsec:comparison}).
The two methods were comparable (with PDE-FIND showing a minor advantage) on noiseless and small noise data, while Algorithm~\ref{alg:2} showed an advantage over PDE-FIND with larger noise data.
We believe the success of Algorithm~\ref{alg:2} in this experiment is due to the smoothing properties of the heat operator, which provides considerable robustness.

In terms of the hyperparameters, the experiments with Algorithm~\ref{alg:1} in Sections~\ref{sec:Parabolic} and~\ref{sec:Wave} show that increasing the height of the model gives better predictability.
We also found in Section~\ref{sec:numerics_Burgers} that one can effectively estimate the viscosity parameter $\nu>0$ at no expense in the error.
These experiments demonstrate a potential for the use of MFVs as features for learning PDEs.
A more systematic comparison of our algorithms with other methods as well as analysis of the effect of hyperparameters is left for future work.

	We conclude by discussing several other directions in which this work could be extended.
\begin{itemize}
\item \textbf{Beyond PDEs.} An important next step is to investigate the use of models as features in learning algorithms in contexts beyond PDEs.
	We believe that natural directions to investigate include analysis of meteorological data~\cite{weather,CHEN201852}, image and remote sensing recognition~\cite{CIFAR-100,RSSCN7}, and applications to fluid dynamics~\cite{fluid_review,ling_kurzawski_templeton_2016}.
Such extensions
would parallel the current use of signatures in data science well beyond the scope of ODEs.

\item \textbf{Universality.} It would be of interest to understand universality properties of models, i.e. in what sense and under which conditions can one approximate general functions of the input $(\{u^{(i)}\}_{i\in\CJ}, \xi)$ with \textit{linear} functions of the model.
Beyond their importance in machine learning, such universality properties are of deep mathematical interest; a celebrated result is that linear functions of the signature (see Definition~\ref{def:sigs}) approximate, uniformly on compact sets, continuous function of rough paths modulo \textit{tree-like equivalence}~\cite{hambly2010uniqueness, BOEDIHARDJO2016720}.

\item \textbf{Further learning algorithms.}	It will be important to explore the utility of `model features' when combined with learning algorithms beyond linear regression (the only tool used in this article), such as with neural networks and random forests.
Similarly, it would be important to \textit{kernelise} the model feature vector efficiently.
This would allow for use of popular kernel learning methods, such as support vector machines,
and of the maximum mean discrepancy (MMD) of~\cite{gretton2012kernel} to compare samples drawn from different distributions.
An MMD from the kernelised signature map was used in~\cite{CO18Sigs_published} to define a metric on the laws of stochastic process indexed by time, and fast signature kernelisation algorithms 
	%which scale linearly with the level (i.e. height)
	were introduced in~\cite{KO19};
	extending these results to models would be of significant interest.

\item \textbf{Higher dimensions.} In order to apply the ideas in this paper to data in high dimensional spaces, it would be important to improve the computation of models.
It took\footnote{On a laptop with 4 Cores (1.4 GHz) and 16 GB memory.} between $0.2$ to $0.5$ seconds
to compute one model in Sections~\ref{sec:Parabolic} and~\ref{sec:Wave}, and approximately $90$ seconds to perform one run of Algorithm~\ref{alg:2} in Section~\ref{sec:numerics_Burgers}.
%$0.49$ seconds to compute one model $\CM^4_{3,1}$ of degree $\leq 5$ for the parabolic equation experiment in Section~\ref{sec:Parabolic}, $0.28$ seconds to compute ${\CM}^4_{2,1}$ with ${\CM}^0_{2,1} = \{I_c[u_0], I_s[v_0]\}$ and degree $\leq 1.5$ for the wave equation experiment in Section~\ref{sec:Wave},
%and it took $582.5$ seconds to perform one run of Algorithm~\ref{alg:2} in Section~\ref{sec:numerics_Burgers}.
The computation time in higher spatial dimensions would be significantly longer.

In this direction, there are a number of works aiming to solve high dimensional PDEs with learning algorithms, such as~\cite{MR3847747,MR3874585}.
Since, with the choice of operator $I$ in our experiments, elements of the models are solutions to special PDEs, it would be interesting to see if these methods could make it feasible to compute the model features 
in high dimensional spaces.

\item \textbf{Operator $I$ hyperparameter.} The operator $I$ in the definition of a model is a hyperparameter which needs to be chosen from a very large space (the infinite-dimensional space of linear operators).
	In our experiments, we mostly used knowledge of the \textit{linear part} of the PDE (heat or wave operators) to choose $I$.
	However, if the PDE is completely unknown, or if the output $u$ does not come from a PDE at all, then one would need a systematic way to choose this hyperparameter.
	The same applies to the other hyperparameters, such as $n,m,\ell,q$, but these take values in a smaller space for which standard hyperparameter tuning (e.g. cross-validation, or sparse linear regression similar to PDE-FIND~\cite{PDE-FIND}) is feasible.
	Note that the problem of choosing $I$ does not arise in the context of signatures simply because one hardcodes $I$ as convolution with the Heaviside step function $J(t)=\bone_{t>0}$.
	We could of course likewise hardcode $I$, e.g., as the inverse of the heat operator~\eqref{eq:SHE},
	but if one believes the output $u$ should behave like the solution to a wave equation, this will likely yield poor performance.
	How to choose $I$ in a general context is therefore an important theoretical and practical question.
	%, which is present, we believe, even in the 1-dimensional case of time-ordered data.
\end{itemize}

%\begin{appendices}

%\smallskip
%\noindent
%\textbf{Author Contribution.} All authors contributed equally.
%
%\smallskip
%\noindent
%\textbf{Conflict of Interest.} There are no identified conflicts of interest.

%\end{appendices}

\noindent
\textbf{Acknowledgements.} The authors would like to thank the anonymous referees for their thorough reading of the paper and suggestions for improvements.

\medskip

\noindent
\textbf{Data availability statement.} All code and data are publicly available at \begin{center}
\href{https://github.com/andrisger/Feature-Engineering-with-Regularity-Structures.git}{https://github.com/andrisger/Feature-Engineering-with-Regularity-Structures.git} 
\end{center}
	
	\bibliographystyle{Martin}
	\bibliography{refs}
	
\end{document}